\documentclass{article}
\usepackage{arxiv}

\usepackage[utf8]{inputenc} 
\usepackage[T1]{fontenc}    
\usepackage{hyperref}       
\usepackage{url}            
\usepackage{booktabs}       
\usepackage{amsfonts}       
\usepackage{nicefrac}       
\usepackage{microtype}      
\usepackage{lipsum} 
\usepackage{float}        
\usepackage{graphicx}
\usepackage{doi}
\usepackage{cite}
\usepackage{amsmath,amssymb}
\usepackage{algorithmic}
\usepackage{textcomp}
\usepackage{subcaption}
\usepackage{booktabs}
\usepackage{multirow}
\usepackage{makecell}
\usepackage{color}
\usepackage{changepage}
\usepackage{bm}
\usepackage{pifont}
\usepackage{bbding}

\title{Optimization of Module Transferability in Single Image Super‑Resolution: Universality Assessment and Cycle Residual Blocks}

\newif\ifuniqueAffiliation
\uniqueAffiliationtrue

\ifuniqueAffiliation 
\author{Haotong Cheng*\\
College of Electronic Science and Engineering\\
	Jilin University\\
	Changchun, China \\
	\texttt{chenght9923@mails.jlu.edu.cn} \\
	\And
	Zhiqi Zhang \\
	College of Computer Science and Technology\\
	Jilin University\\
	Changchun, China\\
	\texttt{zhangzq2023@mails.jlu.edu.cn} \\
	\And
Hao Li \\
College of Electronic Science and Engineering\\
Jilin University\\
Changchun, China \\
\texttt{lihao1723@mails.jlu.edu.cn} \\
	\And
Xinshang Zhang \\
College of Electronic Science and Engineering\\
Jilin University\\
Changchun, China \\
\texttt{zhangxs2222@mails.jlu.edu.cn} \\
}
\else

\author[1]{%
	\hspace{1mm}David S.~Hippocampus\thanks{\texttt{hippo@cs.cranberry-lemon.edu}}}%
	}
	\author[2]{%
Elias D.~Striatum\thanks{\texttt{stariate@ee.mount-sheikh.edu}}}%
}
\affil[1]{Department of Computer Science, Cranberry-Lemon University, Pittsburgh, PA 15213}
\affil[2]{Department of Electrical Engineering, Mount-Sheikh University, Santa Narimana, Levand}
\fi

\renewcommand{\headeright}{} 
\renewcommand{\undertitle}{} 
\renewcommand{\shorttitle}{} 

\hypersetup{
pdftitle={A template for the arxiv style},
pdfsubject={q-bio.NC, q-bio.QM},
pdfauthor={David S.~Hippocampus, Elias D.~Striatum},
pdfkeywords={First keyword, Second keyword, More},
hidelinks, 
}

\begin{document}
\maketitle

\begin{abstract}
	Deep learning has substantially advanced the Single Image Super‑Resolution (SISR). However, existing researches have predominantly focused on raw performance gains, with little attention paid to quantifying the transferability of architectural components. In this paper, we introduce the concept of ``Universality'' and its associated definitions which extend the traditional notion of ``Generalization'' to encompass the modules' ease of transferability. Then we propose the Universality Assessment Equation (UAE), a metric which quantifies how readily a given module could be transplanted across models and reveals the combined influence of multiple existing metrics on transferability. Guided by the UAE results of standard residual blocks and other plug‑and‑play modules, we further design two optimized modules, Cycle Residual Block (CRB) and Depth‑Wise Cycle Residual Block (DCRB). Through comprehensive experiments on natural-scene benchmarks, remote-sensing datasets and other low-level tasks, we demonstrate that networks embedded with the proposed plug-and-play modules outperform several state-of-the-arts, reaching a PSNR enhancement of up to  0.83dB or enabling a 71.3\% reduction in parameters with negligible loss in reconstruction fidelity. Similar optimization approaches could be applied to a broader range of basic modules, offering a new paradigm for the design of plug-and-play modules. 
\end{abstract}

\keywords{Generalization; Optimization;  Super-resolution; Universality; Plug-and-Play}
\section{Introduction}
\label{sec:introduction}
Single Image Super-Resolution (SISR) reconstructs high-resolution images from low-resolution inputs. However, as application scenarios expand, deep SISR methods  increasingly exhibit limited generalization. Therefore, numerous intricately designed architectures have been proposed to enhance model performance across diverse scenarios.

Since Dong et al.\cite{SRCNN} firstly introduced deep learning methods into image super-resolution tasks, deep learning-based SISR methods have gradually become mainstream. Methods such as SRCNN \cite{SRCNN}, FSRCNN (Fast SRCNN) \cite{FSRCNN}, and ESPCN (Efficient Sub-Pixel CNN) \cite{ESPCN} harness the learning capabilities of shallow convolutional networks, leading to notable performance improvements over traditional methods. Subsequently, models such as VDSR (Very Deep CNN for SR) \cite{VDSR}, RED-Net (Residual Encoder-Decoder Network) \cite{RED}, and EDSR (Enhanced Deep SR) \cite{EDSR} began exploring deeper architectures to improve the SISR performance, among which Residual Blocks  (RB) have become a commonly used component. For instance, EDSR cascades several residual modules that remove Batch Normalization (BN) layers to serve as the feature extraction layer. MSRN (Multi-Scale Residual Network) \cite{MSRN} introduces convolution kernels of different sizes within residual modules. Residual blocks and their optimized variants have become one of the fundamental components in SISR networks.

Inspired by DenseNet, several SISR networks based on dense convolution modules are also proposed. SRDenseNet \cite{SRDense} builds the feature inference blocks by cascading dense connection modules and skip connections between different layers. RDN (Residual Dense Network) \cite{RDN} combines residual modules and dense connection modules to form a residual-dense network. MemNet \cite{MemNet} constructs a memory module through dense connections between different convolution layers and gating units to maintain long-term dependencies between features at different layers.

In recent years, various attention modules have been proposed to improve the model performance. For instance, RCAN \cite{RCAN} introduces channel attention into residual modules and builds a dual-layer residual structure to enhance feature inference capabilities. SAN \cite{SAN} proposes second-order channel attention and constructs non-local enhanced residual groups. Additionally, Transformer-based architectures have also demonstrated considerable potential in SISR tasks.  SwinIR (Image Restoration Using Swin Transformer) \cite{SwinIR} achieves better results with fewer parameters by constructing a Residual Swin Transformer module.  HiT-SR \cite{HiT} builds a Hierarchical Transformer module that expands windows to aggregate hierarchical features, enabling feature inference from local regions to long-range dependencies.

\begin{figure}[h]
	\centering
	\subfloat[Structure of IPG \cite{IPG}]{
		\includegraphics[width=0.24\textwidth]{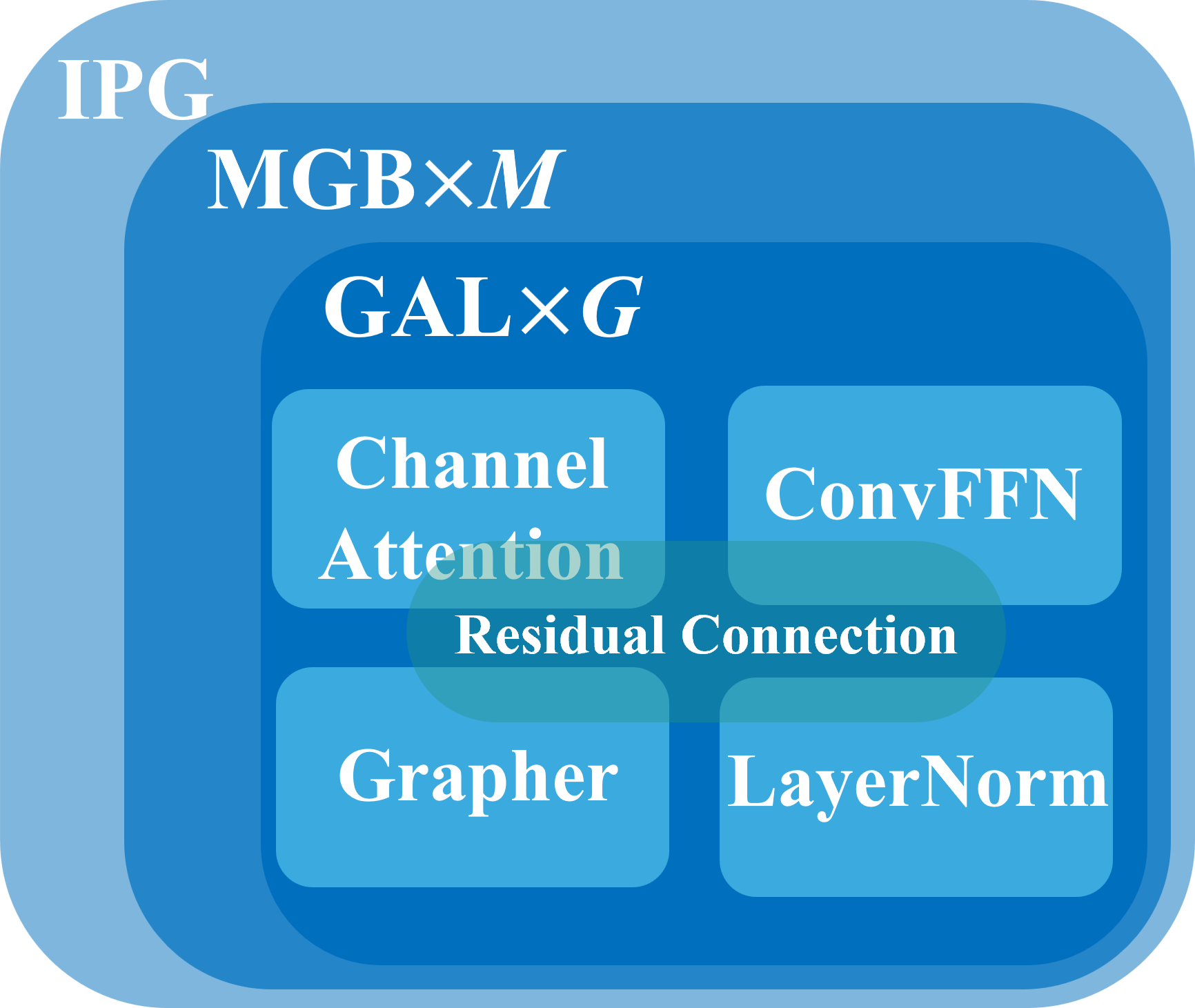}
	}
	\quad\quad
	\subfloat[Structure of SwinIR \cite{SwinIR}]{\includegraphics[width=0.24\textwidth]{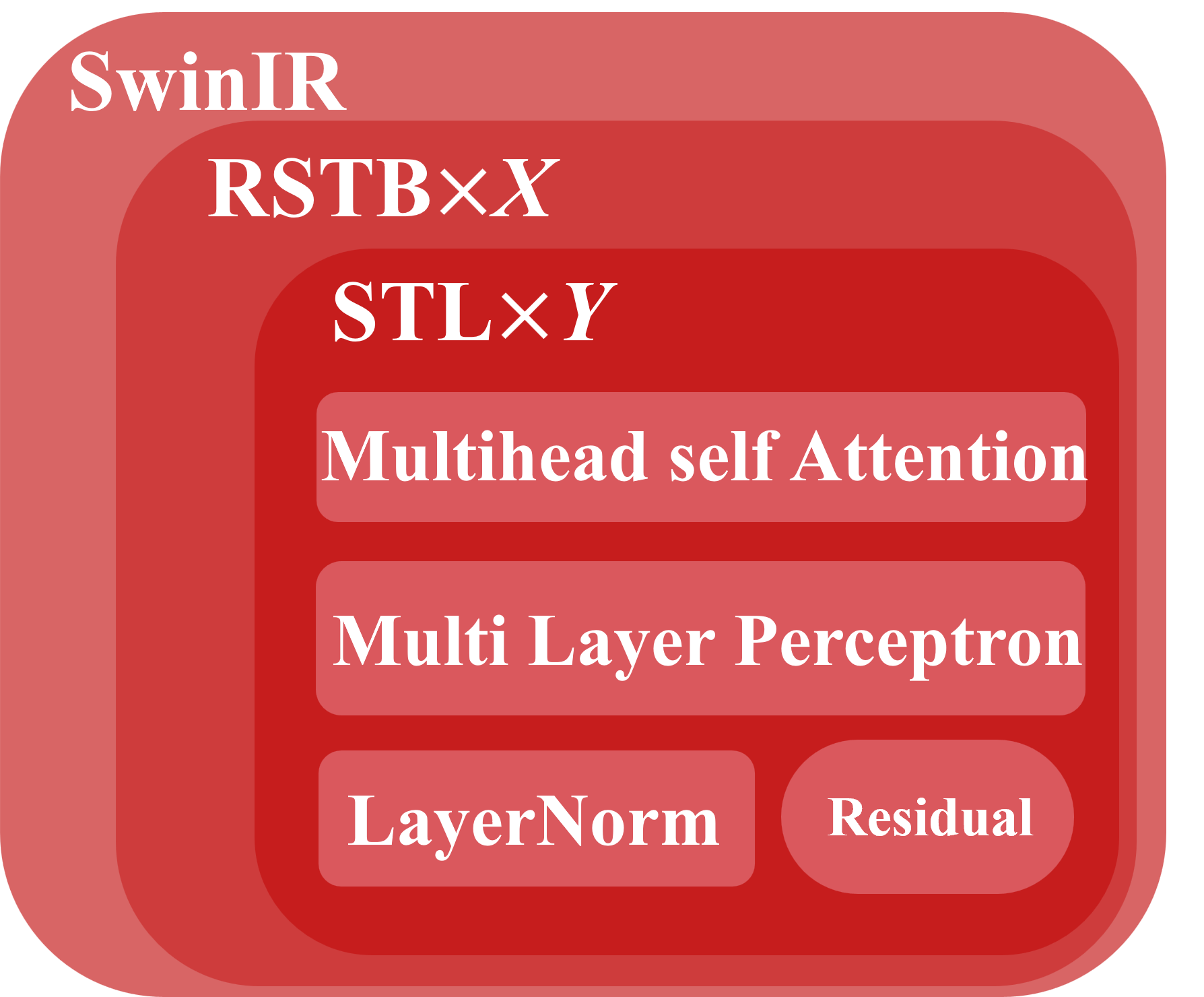}}
	\caption{The highly coupled internal structure of IPG (CVPR 2024) and SwinIR (ICCV 2023).}
	\label{IPG}
\end{figure}

It is evident that various modules form the core components of SISR models, including residual modules, Transformer modules, etc. Additionally, researchers increasingly favor the plug-and-play modules characterized by simplicity in structure, ideal performance and strong transferability. For instance, ConvFFN was initially introduced in ViT \cite{ViT} and is still applied in IPG \cite{IPG}. ResBlock, primarily proposed in \cite{Resnet}, is still widely embedded in various networks to enhance model performance. In existing studies, however, the plug-and-play modules are characterized only qualitatively, and the concept of ``Generalization'' primarily focuses on performance across different datasets. Although numerous state-of-the-arts have markedly improved the SISR performance, their internal modules remain difficult to operate independently. For instance, although IPG \cite{IPG} achieves ideal reconstruction performance using a delicately optimized GNN architecture, its modules are highly coupled--a limitation also observed in SwinIR \cite{SwinIR}, as shown in Fig.\ref{IPG}.  

To enable a quantitative assessment of module transferability and provide a novel perspective on plug-and-play module optimization, this paper proposes the definition of ``Universality'' and its associated concepts. Subsequently, the \textbf{U}niversality \textbf{A}ssessment \textbf{E}quation (UAE) is designed by analyzing the pivotal characteristics of existing plug-and-play modules and other state-of-the-art networks.
\textbf{C}ycle \textbf{R}esidual \textbf{B}locks (CRB) and \textbf{D}epth-wise \textbf{C}ycle \textbf{R}esidual \textbf{B}locks (DCRB) are further derived on the basis of RB under the guidance of UAE results. Experiments demonstrate that models embedded with the optimized modules could achieve a 0.83dB increase in PSNR, or reduce parameters by up to 71.3\% with minimal performance loss. The main contributions of this paper are as follows:

1) The definition of ``Universality'' is proposed. Compared to ``Generalization'', ``Universality'' describes modules' ease of transferability and is quantified by the Universality Assessment Equation (UAE).

2) CRB and DCRB are designed as optimized plug-and-play modules referring to UAE results. The inherent optimization mechanisms of CRB and DCRB are also revealed from the standpoint of back propagation process. 

3) The effectiveness of CRB and DCRB is validated through experiments on various SISR tasks, and the UAE optimization strategy is further verified through evaluation on various low-level tasks such as denoising and deblurring. Experiments demonstrate that our optimization strategy features strong generalizability across different vision tasks, offering a new paradigm for the optimization of plug-and-play modules.

\section{Related Works}
\subsection{Models for Single Image Super-resolution}
Early SISR networks could be broadly classified into four categories based on their backbone models: CNN-based models (\cite{SRCNN,EDSR,VDSR}), GNN-based models (\cite{IPG,GCN,GSR,DSFT}), GAN-based models (\cite{ESRGAN}), and Transformer-based models (\cite{IPT,SwinIR,SRFormer,UFormer}). While highly integrated models could reduce computational costs and improve the performance, they also make it difficult to achieve efficient plug-and-play compatibility across different networks. For instance, the RSTB module in SwinIR \cite{SwinIR} contains STL layers, which in turn incorporate MSA (Multi-Head Self-Attention) and MLP (Multi-Layer Perceptron). 

Recent hybrid architectures introduce new paradigms but compromise transferability as well. SRMamba-T \cite{SRMamba-T} merges Mamba's sequential scanning with Transformer's global attention for long-range dependency modeling. However, the fundamental difference between Mamba's linear scanning mechanism and Transformer's global self-attention necessitates intricate signal pathways, resulting in considerable architectural coupling. HSR-KAN \cite{HSR-KAN} integrates KANs (Kolmogorov-Arnold Networks) with CNN and MLP to reach a trade-off between efficiency and quality , but its specialized fusion structure (e.g., KAN-CAB module) severely limits its task generalizability. Although the BUFF (Bayesian Uncertainty Guided Diffusion Model) \cite{BUFF} introduces Bayesian methods into diffusion models to relax the assumption of independent noises, however, the instability of Bayesian training substantially increases model coupling and computational intensity. Furthermore, the tight coupling in diffusion models between the forward diffusion and reverse denoising processes precludes any individual component from functioning independently. Therefore, while these advances boost performance, their intricate integrations impede module reuse across architectures.

\subsection{Plug-and-Play Modules in SISR Networks}
The concept of plug-and-play modules emerges from the recognition that modular components could improve reproducibility across different SISR architectures. Early successful examples include the widespread use of RB across diverse networks, from CNN-based EDSR \cite{EDSR} to GAN-based ESRGAN \cite{ESRGAN}, demonstrating the value of structurally simple yet effective components.  

The attention mechanisms represent one of the most successful transferable modules in recent years. Beyond RCAB's \cite{RCAN} channel attention, researchers developed increasingly sophisticated attention variants. For instance, MDAB (Multi-scale Dilated Attention Block) \cite{MDAB} offers a lightweight attention mechanism. However, its effectiveness critically depends on the pre-processing by  a dedicated LRM (Local Residual Module). Consequently, transplanting MDAB without its coupled LRM generally leads to performance degradation. 

These plug-and-play modules significantly enhance model performance but remain relatively scarce and challenging to design, since they demand highly streamlined structures while preserving its ideal performance. Therefore, analyzing the module structures from a universal perspective is crucial for designing improved plug-and-play modules.

\subsection{Generalization and Universality}
``Generalization'' refers to a model's capacity to maintain high performance when evaluated on data distinct from its training distribution \cite{Generalization}. Existing researches exploring ``universality'' predominantly focus on the developments of unified frameworks and task formulations. For instance, IPT \cite{IPT} leverages large-scale pre-training to unify diverse low-level vision tasks, while DGUNet \cite{DGUNet} proposes an interpretable, unified deep neural network architecture. Similarly, Uni-COAL\cite{Uni-COAL} introduces a unified framework for Magnetic Resonance Imaging super-resolution. Although these approaches enhance model performance by improving framework universality, they have not thoroughly investigated or provided quantitative evaluation for the module transferability. Therefore, this paper proposes the concept of universality to characterize the transferability of individual components. This quantification facilitates the systematic design of plug-and-play modules aimed at enhancing model generalization and framework universality.

\section{Proposed Methods}
In this section, definitions for ``Universality'' and its associated concepts are proposed to lay the foundation for a quantitative description of module transferability, namely the Universality Assessment Equation. We further optimize RB into CRB/DCRB through UAE analysis and the optimization mechanism under UAE guidance is elucidated from the perspective of back propagation. Finally, a generalized designing principle for CRB is explained to provide a clearer picture on module optimization. 

\subsection{``Universality'' and ``Positive  Universality''}
Let $M_i$ and $M_k$ denote two network architectures ($i\neq k$), and let $B_j$ be a module originally embedded in $M_i$. We define $B_j$  to be universal for $M_k$ if it could be incorporated into $M_k$ without  significant structural modifications while preserving its intended functionality. Thus, universality measures modules' ease of transferability across architectures. By contrast, generalization refers to a model’s ability to maintain an ideal performance when evaluated on datasets distinct from its training set. A summary of these distinctions is presented in Table \ref{Differences}.
\begin{table}[h]
	\centering
	\caption{Differences between ``Universality'' and ``Generalization''.}
	\begin{tabular}{cccc}
		\toprule
		\textbf{Concept} & \textbf{Subject} & \textbf{Object} & \textbf{Evaluation Criteria} \\
		\midrule
		\multirow{2}{*}{Universality}&\multirow{2}{*}{\makecell{Local\\ module}} & \multirow{2}{*}{\makecell{Complete\\ model}}& \multirow{2}{*}{\makecell{The ease of  \\module transferability}}\\ 
		& & & \\
		\midrule
		\multirow{2}{*}{Generalization}&\multirow{2}{*}{\makecell{Complete \\model}} & \multirow{2}{*}{\makecell{Brand new \\test data}}& \multirow{2}{*}{\makecell{Whether model \\performance remains ideal}}\\		
		&&&\\
		\bottomrule
	\end{tabular}
	\label{Differences}
\end{table}

Although universality could describe a module’s intrinsic transferability, it is difficult to characterize its precise impacts on model performance after integration, where directly embedding a plug‑and‑play module may degrade the performance. For instance, inserting a PnP denoising block \cite{PnP} into the EDSR backbone is likely to trigger performance degradation, because the denoiser's learned feature distribution breaks the identity mappings that deep residual blocks rely on, which perturbs the signals. Therefore, we propose the ``Positive Universality (PU)'' to describe that module $B_j$ could effectively improve the performance of a brand new model after being detached from its parent model. The opposite concept is ``Negative Universality (NU)''. Our experiments validate that the CRB is such a module possessing  PU property.

Universality and generalization are not completely isolated. Differences between the previously proposed plug-and-play modules (\cite{Resnet,RCAN,2025 CVPR plug-and-play 1,2025 CVPR plug-and-play 2}) reveal that while more complex blocks could achieve superior generalizability, they also typically incur larger parameter counts and more elaborate computational graphs, all of which may deteriorate the ease of transferability. Therefore,  generalizability could be enhanced by appropriately decreasing the module universality.  

\subsection{Universality Assessment Equation (UAE)}
Through the structural differences between plug-and-play modules (RB \cite{Resnet}, RCAB \cite{RCAN}, ConvFFN \cite{ViT}, etc.) and non-plug-and-play modules (RSTB \cite{SwinIR}, GAL \cite{IPG}, etc.), we identify four shared factors that influence most on the universality, namely ``The nesting level of other blocks'', ``Total number of parameters'', ``Cascading of forward propagation layers'' and ``Input feature dimensions''. Six UAE forms are chosen in Experimental Analysis to validate the conclusions.

Generally, transferability difficulty rises with an increased module nesting level due to stronger internal coupling (e.g., GAL \cite{IPG}). Parameter‑scarce modules are sensitive to parameter growth until a threshold, beyond which further increases yield diminishing returns and a risk of overfitting (e.g., standard convolution modules \cite{SRCNN} and RB \cite{Resnet}). Augmenting the number of forward propagation layers increases computational overhead and correspondingly deteriorates module universality. However, this degradation is less pronounced than the increases contributed by nested architectures, since the nested modules already incorporate multiple propagation stages (e.g., ConvFFN \cite{ViT} and RSTB \cite{SwinIR}). Consequently, we compute the UAE per input feature‑map unit, defining it abstractly as: 
\begin{equation}
	\phi=\frac{\alpha(l)\times\beta(k)\times \theta(n)}{f},\ 
	\begin{cases}
		\theta''(n)<0, \\
		\beta''(k)>\alpha''(l)>0.
	\end{cases}
	\label{UAE abstract}
\end{equation}
where $k$ is the number of nested sub-modules, $n$ the total parameters, $l$ the forward propagation layers, $f$ the input feature units. $\phi$ is inversely related to universality. $\theta''(n)<0$ denotes a diminishing sensitivity to parameter increases and a slowing universality decline, while $\beta''(k)>\alpha''(l)>0$ indicates that nesting modules erode universality more rapidly than cascaded layers.

Although the specific forms of $\alpha(l),\beta(k)$ and $\theta(n)$ vary, the choice of UAE does not alter the relative ranking of module universality. Eq.\eqref{UAE example} provides one such instance, and five more UAE variants —despite spanning from $10^{-1}$ to $10^{4}$—yield identical module orderings. Consequently, modules with UAE values close to the baseline exhibit similar transferability, and lower UAE results generally indicate better universality. 
\begin{equation}
	\phi=\frac{l \times e^{k+1} \times \lg \left(\frac{n}{100}\right)}{f}
	\label{UAE example}
\end{equation}
\begin{figure*}[t]
	\centering
\includegraphics[width=0.55\textwidth]{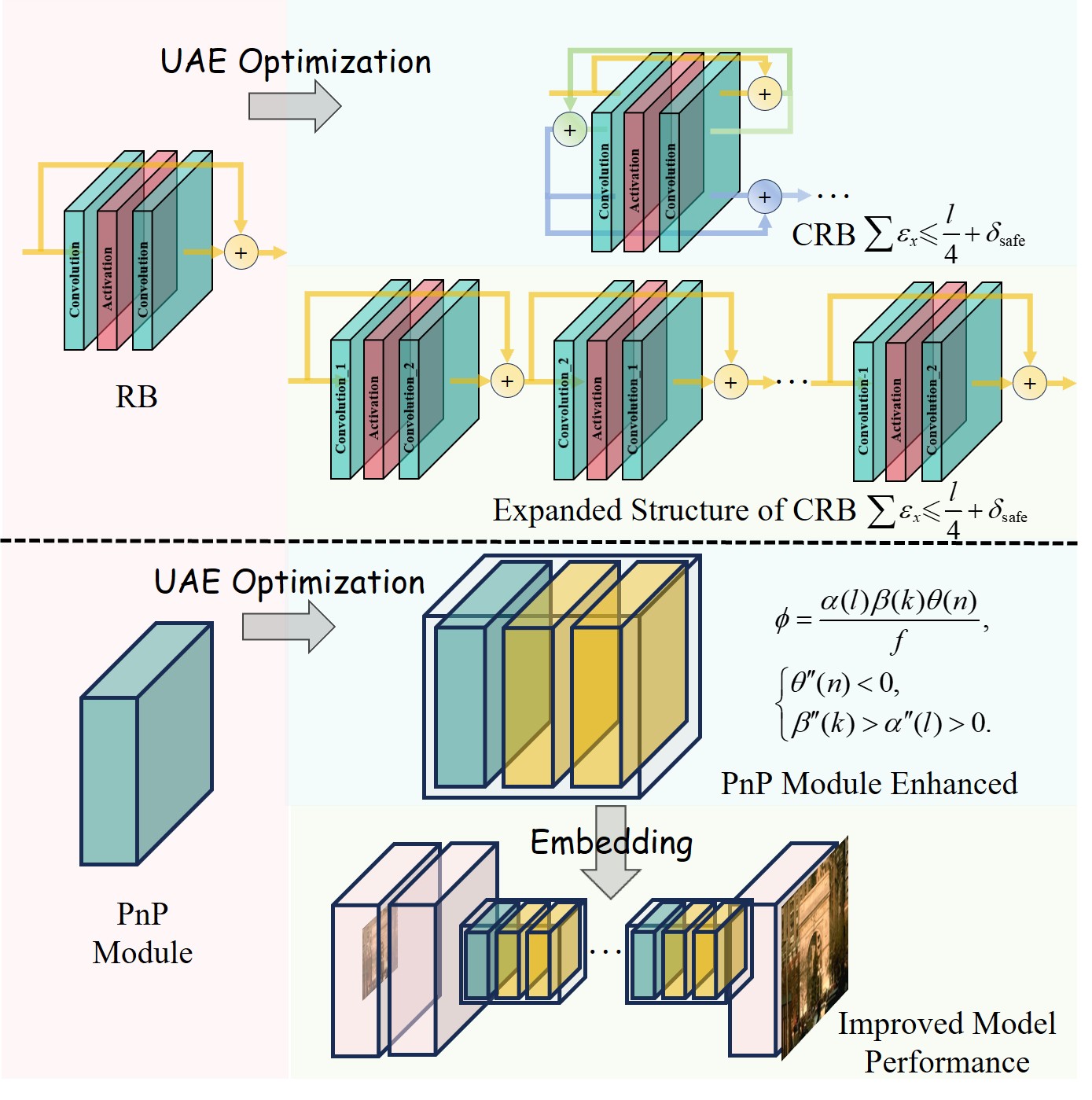}
	\caption{Overview of the proposed blocks and the UAE optimization strategy.}
	\label{Optimized blocks}
\end{figure*}

\subsection{Cycle Residual Block and Depth-wise Separable Cycle Residual Block}
According to Eq.\eqref{UAE abstract}, RB could be optimized from three perspectives: $k$, $l$, and $n$. This study optimizes RB with respect to $l$ and $n$, yielding CRB and DCRB respectively, as  illustrated in Fig.\ref{Optimized blocks}. Specifically, CRB reuses RB's standard convolution layers, thereby improving the layer-utilization efficiency without adding any extra parameter. DCRB employs depth-wise separable convolution to reduce the parameters, resulting in a more lightweight and integrated module structure.

Let $X\in\mathbb{R}^{B\times C\times H\times W}$ denote the input tensor, where $B$, $C$, $H$ and $W$ correspond to  batch size, channel dimensions, height and width, respectively. The CRB and its depth-wise variant DCRB are then defined as Eq.\eqref{CRB details}, where Conv($\cdot$) represents standard convolution for CRB and depth-wise separable convolution for DCRB.
\begin{equation}
	\begin{cases}
		X_{\text{mid}}=X_{\text{input}}+\text{Conv}(\text{ReLU}(\text{Conv}(X_{\text{input}})))\\Y_{\text{output}}=X_{\text{mid}}+\text{Conv}(\text{ReLU}(\text{Conv}(X_{\text{mid}})))
	\end{cases}
	\label{CRB details}
\end{equation}

To elucidate the optimization mechanism of CRB, we further examine its back propagation process. Initially, to assume that gradient magnitude could be taken as a approximation for convergence rate, we enforce several prior conditions of RB and CRB to be identical, all of which are ensured in our experiments via the following configurations: (1) We use identical hyperparameters (learning rate, etc.), (2) The BatchNorm layers of RB/CRB are removed, and the global random seeds are fixed to control the identical condition number $\kappa=L/\mu$, (3) We apply the same dropout rate and other settings to ensure similar local Hessian structures. Under these controlled conditions, the observed difference in gradient magnitude reflects the difference of per-step update, which could be interpreted as the convergence speed.

Let $y=f_L(f_{L-1}(\cdots f_l(x)\cdots))$ denote a neural network where $x$ is the input of layer $f_l(\cdot)$ and $y$ the model output, the back propagation is then given by Eq.\eqref{back propagation}.
\begin{equation}
	\frac{\partial L(x)}{\partial x} = \frac{\partial L(x)}{\partial y}\cdot\frac{\partial y}{\partial x}=\frac{\partial L(x)}{\partial y} \cdot\prod_{i=l}^{L-1}\frac{\partial x_{i+1}}{\partial x_i}=\frac{\partial L}{\partial y}\cdot \prod_{i=l}^{L-1}J_i
	\label{back propagation}
\end{equation}
where $L(x)$ stands for losses and $J_i$ the Jacobian matrix. Let $\mathbf{F}_l(\cdot)$ denote the nonlinear mapping (e.g., convolution followed by activation ) in the $l$-th residual block. Since each block implements $x_{l+1}=x_l+\mathbf{F}_l(x_l)$, the total derivative of the loss with respect to initial input $x_0$ follows by the repeated application of chain rule:
\begin{equation}
	\frac{\partial L(x)}{\partial x_0}=\frac{\partial L(x)}{\partial x_n}\cdot\prod_{l=0}^{n-1}(\mathbf{E}+\frac{\partial \mathbf{F}_l(x_l)}{\partial x_l})=\frac{\partial L(x)}{\partial x_n}\prod_{l=0}^{n-1}J_l^{\text{RB}}
	\label{RB gradient}
\end{equation}
where $\textbf{E}\in\mathbb{R}^{d\times d}$ is the identity matrix. From Eq.\eqref{CRB details}, CRB could be expressed as :
\begin{equation}
	y = x+\mathbf{F}(x) + \mathbf{F}(x+\mathbf{F}(x))
\end{equation}

By conducting similar calculations,  we derive the  back propagation of a model made up of $n$ CRBs:
\begin{equation}
	\begin{aligned}
		\frac{\partial L(x)}{\partial x_0}=\frac{\partial L(x)}{\partial x_n}\prod_{l=0}^{n-1}\Big[ (\mathbf{E}+\frac{\partial \mathbf{F}_l(x_l)}{\partial x_l})(\mathbf{E}+\frac{\partial \mathbf{F}_l(y_l)}{\partial y_l})\Big] 
		=\frac{\partial L(x)}{\partial x_n}\prod_{l=0}^{n-1}J_l^{\text{CRB}}.
	\end{aligned}
	\label{CRB gradient}
\end{equation}
where $y_l=x_l+\mathbf{F}_l(x_l)$. Considering that $\mathbf{E}+\partial \mathbf{F}_l(x_l)/\partial x_l$ are matrices, we explicitly employ the spectral norm $\Vert\cdot\Vert_2$ to characterize the gradient amplification. Firstly, it is proved that when $A$ and $B$ are symmetric positive semidefinite matrices and commute with the identity matrix, Eq.\eqref{proof} is valid.
\begin{equation}
	\Vert(E+A)(E+B)\Vert_2>\Vert(E+A)\Vert_2.
	\label{proof}
\end{equation}

Let $A, B\in\mathbb{R}^{d\times d}$ satisfy: (1) ``$A$ and $B$ are symmetric positive semidefinite matrices'', (2) ``$A$ and $B$ commute, i.e. $AB=BA$''. Thus, there exists orthogonal matrix $Q$ satisfying Eq.\eqref{orthogonal}.
\begin{equation}
	\begin{aligned}
		Q^\top AQ=\text{diag}(\alpha_1,\cdots,\alpha_d),Q^\top BQ=\text{diag}(\beta_1,\cdots,\beta_d).
	\end{aligned}
	\label{orthogonal}
\end{equation}
where $\alpha_i,\ \beta_i>0$. The matrices $E+A$ and $E+B$ are diagonalized under this basis as Eq.\eqref{diagonalized}.
\begin{equation}
	\begin{aligned}
		Q^\top (E+A)Q=\text{diag}(1+\alpha_1,\cdots,1+\alpha_d),Q^\top (E+B)Q=\text{diag}(1+\beta_1,\cdots,1+\beta_d).
	\end{aligned}
	\label{diagonalized}
\end{equation}

Thus, we have $(E+A)(E+B)$ in this basis equal to $\text{diag}((1+\alpha_1)(1+\beta_1),\cdots,(1+\alpha_d)(1+\beta_d))$. By the definition of $\Vert\cdot\Vert_2$, we derive Eq.\eqref{spectral norm}.
\begin{equation}
	\begin{aligned}
		\Vert E+A\Vert_2&=\underset{i}{\max}(1+\alpha_i), \Vert (E+A)(E+B)\Vert_2&=\underset{i}{\max}[(1+\alpha_i)(1+\beta_i)].
	\end{aligned}
	\label{spectral norm}
\end{equation}
where $\underset{i}{\max}\beta_i\triangleq\beta_{\max}\geq0$. Finally we arrive at the expression given in Eq.\eqref{finally}.
\begin{equation}
	\begin{aligned}
		\Vert(E+A)(E+B)\Vert_2&=(1+\alpha_{\max})(1+\beta_{\max}) >(1+\alpha_{\max})&=\Vert(E+A)\Vert_2
	\end{aligned}
	\label{finally}
\end{equation}

Although the proof relies on strong linear algebra assumptions where $J_l^{\text{RB}}$, $J_l^{\text{CRB}}$ may not be symmetric, and its eigenvectors may not be identical, the core insight still holds: the residual structure inherently inserts a gain factor $\geq1$ into the chain product, thereby revealing how residual structures mitigate gradient vanishing/explosion.

Such approximation is grounded in existing physical principles. From the perspective of Neural-ODEs \cite{Neural ODE}, a residual block implements one step of the Euler method: $x(t+\Delta t)=x(t)+\Delta t\cdot f(x(t))$, yielding $F(x)=x(t+\Delta t)-x(t)\approx \Delta t\cdot f(x)$ and $F'\xrightarrow{}0$. Treating $F'$ as a ``small perturbation'' aligns with the fact that residual learning ultimately guides deep modules degenerate into identity function. 

Therefore, even relaxing the symmetric/commuting conditions,  Eq.\eqref{finally} together with Eq.\eqref{CRB gradient} indicate that every added residual branch contributes at least a unit of spectral gain$\geq1$ into the back propagation, leading to smoother gradient flow, faster convergence, and better module performance.

\begin{table*}[t]
	\centering
	\caption{UAE calculation on different modules, where $\text{UAE}_i$ represents six specific forms of UAE and $f=64$. Structure of CRB is chosen as $\varepsilon=2, l=8$.}
	\begin{tabular}{ccccccccc}
		\toprule
		& & RB (baseline) \cite{Resnet} & RCAB \cite{RCAN} & ConvFFN \cite{ViT} & RSTB \cite{SwinIR} & GAL \cite{IPG} & DCRB & CRB \\
		\midrule
		\multirow{3}{*}{UAE Variables} & $k$ & 0 & 1 & 1 & 3 & 3 & 1 & 0 \\
		& $n$ & 73,856 & 148,292 & 17,856 & 86,784 & 56,132 & 1,280 & 73,856 \\
		& $l$ & 4 & 15 & 6 & 11 & 21 & 8 & 8 \\
		\midrule
		$\text{UAE}_1 (\times 1)$ & $\phi_1$ & 0.49 & 5.50 & 1.56 & 27.57 & 49.25 & 1.02 & 0.98 \\
		
		$\text{UAE}_2 (\times 1)$ & $\phi_2$ &  0.17 & 1.73 & 0.69 & 9.38 & 17.92 & 0.92 & 0.34 \\
		
		$\text{UAE}_3 (\times 1)$ & $\phi_3$ & 0.13 & 1.31 & 0.37 & 1.98 & 3.54 & 0.24 & 0.26 \\
		
		$\text{UAE}_4 (\times 1)$ & $\phi_4$ &  1.95 & 82.38 & 9.36 & 303.32 & 1034.29 & 8.18 & 7.80 \\
		
		$\text{UAE}_5 (\times 1)$ & $\phi_5$ & 72.19 & 8159.08 & 463.54 & 25832.26 & 75268.47 & 278.18 & 276.01 \\
		$\text{UAE}_6 (\times 1)$ & $\phi_6$ & 0.08 & 0.71 & 0.29 & 1.20 & 2.30 & 0.38 & 0.16 \\
		\bottomrule
	\end{tabular}
	\label{Universality Assessment}
\end{table*}
\begin{table*}[t]
	\centering
	\caption{Quantitative comparison results (PSNR/SSIM) between several universal modules with $4\times$ SR scale, trained on DIV2K.}
	\begin{tabular}{cccccc}
		\toprule
		Modules & RB (baseline) \cite{Resnet} & RCAB \cite{RCAN} & ConvFFN \cite{ViT} & CRB & DCRB  \\
		\midrule
		$\phi_2$ & 0.17 & 1.73 & 0.69 & 0.34 & 0.92 \\
		
		Set5 \cite{Set5}  & 31.29/0.8632 & 31.17/0.8542 & \textbf{31.71}/\textbf{0.8954} & \underline{31.33}/\underline{0.8648 }& 30.91/0.8538 \\
		Set14 \cite{Set14}& \underline{28.06}/0.6961 & 27.07/\textbf{0.7089} & 27.86/\underline{0.7088} & \textbf{28.49}/0.6971 & 26.69/0.6907 \\
		B100 \cite{B100}& \underline{25.95}/\underline{0.8122} & 25.89/0.8082 &25.91/0.8109 & \textbf{25.97}/\textbf{0.8137} & 25.89/0.8076 \\
		Urban100 \cite{Urban100}& \textbf{23.53}/\underline{0.7597} & 23.49/0.7414 &23.45/0.7502 &\underline{23.52}/\textbf{0.7615} & 23.13/0.7312 \\
		\bottomrule 
	\end{tabular}
	\label{performance-plug-and-play}
\end{table*}
\begin{table*}[t]
	\centering
	\caption{Quantitative comparison (PSNR/SSIM/LPIPS) for RB \cite{EDSR} and CRB/DCRB, using EDSR \cite{EDSR} as the backbone. Ours\textsubscript{$i$} represents the incorporation of corresponding module into the baseline.   }
	\begin{tabular}{cccccc}
		\toprule
		SR Scale & Model & Set5\cite{Set5} & Set14 \cite{Set14} & B100 \cite{B100} & Urban100 \cite{Urban100} \\
		\midrule
		\multirow{3}{*}{$\times 3$} & EDSR (RB) & 33.69/0.8844/0.0707 & 29.87/0.7101/0.0634 & 28.31/0.8540/0.0980 & -/-/- \\
		&Ours\textsubscript{1}(CRB) & \textbf{33.76}/\textbf{0.8846}/\textbf{0.0684} & \textbf{29.92}/\textbf{0.7114}/\textbf{0.0624} & \textbf{28.33}/\textbf{0.8553}/\textbf{0.0972} & -/-/- \\
		&Ours\textsubscript{2}(DCRB) & 32.98/0.8756/0.1248 & 28.38/0.7072/0.0918 & 27.56/0.8421/0.1538 & -/-/- \\
		\midrule
		\multirow{3}{*}{$\times 4$} & EDSR (RB) & 31.29/0.8632/\textbf{0.1092} & 28.06/0.6961/\textbf{0.1085} & 25.95/0.8122/0.2026 & \textbf{23.53}/0.7597/0.0247 \\
		& Ours\textsubscript{1}(CRB) & \textbf{31.32}/\textbf{0.8648}/0.1112 & \textbf{28.49}/\textbf{0.6971}/0.1110 & \textbf{25.97}/\textbf{0.8137}/\textbf{0.2008} & 23.52/\textbf{0.7615}/0.0243 \\
		& Ours\textsubscript{2}(DCRB) & 30.91/0.8538/0.2017 & 26.69/0.6907/0.1690 & 25.89/0.8076/0.2853 & 23.13/0.7312/\textbf{0.0236} \\
		\bottomrule
	\end{tabular}
	\label{multiple scale}
\end{table*}
\subsection{Generalized Forms for CRBs and its Designing Principles}
Although Eq.\eqref{finally} demonstrates that introducing extra residual connections accelerates module convergence without adding extra parameters and is beneficial for performance improvements, we derive an upper bound on the residual number $\varepsilon$ ($\varepsilon\leqslant l/4 + \sum \varepsilon_x\leqslant l/4 + \delta_{\text{safe}}, \delta_{\text{safe}}\geq 0$), where $l$ is the UAE variable inferred from computational graphs, $\delta_{\text{safe}}$ is correlated with hyperparameters and experimental settings, and $\varepsilon_x$ represents the extra residual connections introduced by inputs. Additionally, we theoretically demonstrate that $\sum\varepsilon_x$ is inversely proportional to the module's stability. When $\sum\varepsilon_x>\delta_{\text{safe}}$, the module undergoes gradient explosion. In our ablation study, we reveal the nature of this phenomenon and clarify why the module performance degrades. 

Consider a CRB* as illustrated in Eq.\eqref{new CRB}, whose $l/4=4, \sum\varepsilon_x=|m|+|n|+|z|$. Similar to Eq.\eqref{CRB gradient}, we derive the back propagation  of a model made up of $n$ CRB*s, as shown in Eq.\eqref{new CRB back}.
\begin{equation}
	y_1 = x + \mathbf{F}_1, y_2 = y_1 + \mathbf{F}_2 + mx, y_3 = y_2 + \mathbf{F}_3 + nx, y_4 = y_3 + \mathbf{F}_4 + zx
	\label{new CRB}
\end{equation}
\begin{equation}
	\begin{aligned}
	\frac{\partial L(x)}{\partial x_0}&=\frac{\partial L(x)}{\partial x_n}\prod_{l=0}^{n-1}\Big[\prod_{i=1}^{4}(\mathbf{E}+\frac{\partial \mathbf{F}_{i,l}}{\partial y_{i-1,l}})+m\prod_{i=3}^{4}(\mathbf{E}+\frac{\partial \mathbf{F}_{i,l}}{\partial y_{i-1,l}})+n(\mathbf{E}+\frac{\partial \mathbf{F}_{4,l}}{\partial y_{3,l}})+z\mathbf{E}\Big] \\
&=		\frac{\partial L(x)}{\partial x_n}\prod_{l=0}^{n-1}J_l^{\text{CRB*}}
	\end{aligned}
	\label{new CRB back}
\end{equation}

It is obvious that $1+\sum\varepsilon_x<\Vert J_l^{\text{CRB*}}\Vert_2< 2+\sum\varepsilon_x$, where $\sum\varepsilon_x=|m|+|n|+|z|$. To reveal the threshold conditions that prevent modules from experiencing gradient explosion, we consider an extreme condition where $\sum\varepsilon_x=\delta_{\text{safe}}$, leading to: $\Vert \partial L(x) / \partial x_0\Vert_2 \in \Big((1+\delta_{\text{safe}})^n, (2+\delta_{\text{safe}})^n\Big)$. Therefore, to minimize the risk of gradient explosion, it is necessary to set $\delta_{\text{safe}}=0$. When $\delta_{\text{safe}}>0$, various factors (e.g., hyperparameters) collectively govern the module's convergence, and the risk of gradient explosion escalates as $\delta_{\text{safe}}$ increases.

To generalize our conclusion, we further consider a CRB with $l$ forward propagation layers where $\varepsilon$ residual connections are introduced, as shown in Eq.\eqref{generalized CRB}. 
\begin{equation}
		y_1=x+\mathbf{F}_1, y_2=y_1+\mathbf{F}_2 + c_1x, y_3=y_2+\mathbf{F}_3 + c_2x,\cdots,\ y_\varepsilon=y_{\varepsilon-1}+\mathbf{F}_\varepsilon + c_{\varepsilon-1}x
	\label{generalized CRB}
\end{equation}

Referring to Eqs.\eqref{new CRB} and \eqref{new CRB back}, we derive the back propagation process  for the model made up of $L$ generalized CRBs. 

By defining $D_k=\partial y_k / \partial x$, the recurrence relation for $D_k$ is established as Eq.\eqref{recurrence}.
\begin{equation}
	D_k = (\mathbf{E}+\frac{\partial \mathbf{F}_k}{\partial y_{k-1}})D_{k-1}+c_{k-1}\mathbf{E},k\geq 2
	\label{recurrence}
\end{equation}

By calculating Eq.\eqref{recurrence} , the general form of  $D_k$ goes as:
\begin{equation}
D_k = \prod_{i=1}^{k}(\mathbf{E}+\frac{\partial \mathbf{F}_i}{\partial y_{i-1}})+\sum_{j=1}^{k-1}c_j\prod_{i=j+2}^{k}(\mathbf{E}+\frac{\partial\mathbf{F}_i}{\partial y_{i-1}})+c_{k-1}\mathbf{E}
\end{equation}
When setting $k=l$, we derive the Jacobian matrix of generalized CRB:
\begin{equation}
J_l=D_{k=l}=\prod_{i=1}^{l}(\mathbf{E}+\frac{\partial \mathbf{F}_i}{\partial y_{i-1}})+\sum_{j=1}^{l-1}c_j\prod_{i=j+2}^{l}(\mathbf{E}+\frac{\partial\mathbf{F}_i}{\partial y_{i-1}})+c_{l-1}\mathbf{E}
\end{equation}

Similarly, $1+\sum_{j=1}^{l-1}|c_j|=1+\sum\varepsilon_x<\Vert J\Vert_2<2+\sum\varepsilon_x$. Consequently, to most straightforwardly ensure that CRB could enhance performance with the least probability of encountering gradient explosion, $\delta_{\text{safe}}$ can be simply set to zero.

\begin{table}[t]
	\centering
	\caption{Quantitative results on Potsdam remote sensing dataset \cite{potsdam} under $\times 4$ SR scale. Hyper-parameter settings of IPG and SRFormer are scaled down to accommodate limited computational resources and accelerate training, while EDSR and SRResnet--both featuring moderate parameter counts--retain their full hyper-parameter configurations. The first 700 images of the dataset are used for model training, while the remaining 300 images for performance evaluation.}
	\begin{tabular}{cccc}
		\toprule
		Models & PSNR (dB) & SSIM & LPIPS \\
		\midrule
		SRCNN (ECCV 2014)\cite{SRCNN}  & 29.61 & 0.8572 & 0.2573 \\
		SRResnet (CVPR 2017) \cite{Resnet} & 31.75 & 0.8986 & 0.0606 \\
		RCAN (ECCV 2018) \cite{RCAN} & 31.98 & 0.9019 & 0.0593 \\
		EDSR (CVPR 2017) \cite{EDSR} & 32.18 & 0.9042 & 0.0581 \\
		SRFormer (ICCV 2023)\cite{SRFormer}& 32.24 & 0.9054 & 0.0548 \\
		IPG (CVPR 2024)\cite{IPG}& 32.43 & 0.9058 & \underline{0.0516} \\
		SRFormer (with CRB) & 32.48 & \underline{0.9062} & 0.0537 \\
		EDSR (with CRB)  & 32.49 & \textbf{0.9097} & 0.0526 \\
		EDSR (with DCRB)  & 32.33 & 0.8888 & 0.0543 \\ 
		SRFormer (with DCRB) & 29.40 & 0.9041 & 0.0722 \\
		IPG (with CRB) & \textbf{32.96}& 0.9049 & \textbf{0.0512} \\
		IPG (with DCRB) & \underline{32.85} & 0.9029 & 0.0565 \\
		\bottomrule
	\end{tabular}
	\label{remote sensing}
\end{table}

\section{Experimental analysis}
\subsection{Experimental Setup and Implementation details}
All experiments are conducted on an RTX 4090D GPU using the PyTorch framework. Training is performed on the DIV2K\cite{DIV2K}, BelT\cite{BelT}, Potsdam\cite{potsdam}, with evaluation on B100, Set14, Set5, and Urban100 \cite{B100,Set14,Set5,Urban100}. Reconstruction fidelity is quantified by PSNR and SSIM \cite{PSNR/SSIM}, while perceptual quality is assessed via LPIPS \cite{LPIPS}, as shown in Eq.\eqref{matrics}.
\begin{equation}
	\begin{cases}
		\text{PSNR}(x,y)=10\log_{10}(\frac{\text{MAX}^2}{\text{MSE}(x,y)}) \\
		MSE(x,y)=\frac{1}{MN}\sum_{i=0}^{M-1}\sum_{j=0}^{N-1}[x(i,j)-y(i,j)]^2\\
		SSIM(x,y)=\frac{(2\mu_x\mu_y+C_1)(2\sigma_{xy}+C_2)}{(\mu_x^2+\mu_y^2+C_1)(\sigma_x^2+\sigma_y^2+C_2)}\\
		LPIPS(I_1,I_2)=\sum_l\lambda_l\Vert\phi_l(I_1)-\phi_l(I_2)\Vert_2
	\end{cases}
	\label{matrics}
\end{equation}
where $\text{MAX}$ is the maximum pixel value, $\text{MSE}$ is the mean squared error between the ground truth and reconstructed image. $\mu_1$ and $\mu_2$ are the local mean intensities of the ground‑truth and reconstructed images, $\sigma_1^2$ and  $\sigma_2^2$  represent their corresponding local variances. $\phi_l$ extracts the feature maps at the $l$-th layer of a pretrained network, and $\lambda_l$ denotes the learned weights. During the training of EDSR and the proposed variants, we adopt the $\mathcal{L}_1$ loss and  ADAM optimizer ($\beta_1=0.9$,  $\beta_2=0.999$). The initial learning rate is set to $10^{-5}$. 

When retraining baseline IPG \cite{IPG} and SRFormer \cite{SRFormer}, we adjust hyperparameters to accelerate convergence. Specifically, the number of iterations is reduced to 10,000, the MLP ratio is lowered from 4 to 2, IPG’s embedding dimension is reduced  to 20 with the number of heads set to 4 accordingly, and SRFormer's dimension is set to 64. The core architectures of models, however, are still fully preserved, ensuring that the designing characteristics of networks remain unaffected despite the reductions in parameter size. 

\subsection{Universality Assessment for Diverse Modules}
\begin{figure*}[t]
	\centering
	\begin{minipage}[t]{0.33\textwidth}
		\centering
		\includegraphics[width=\textwidth]{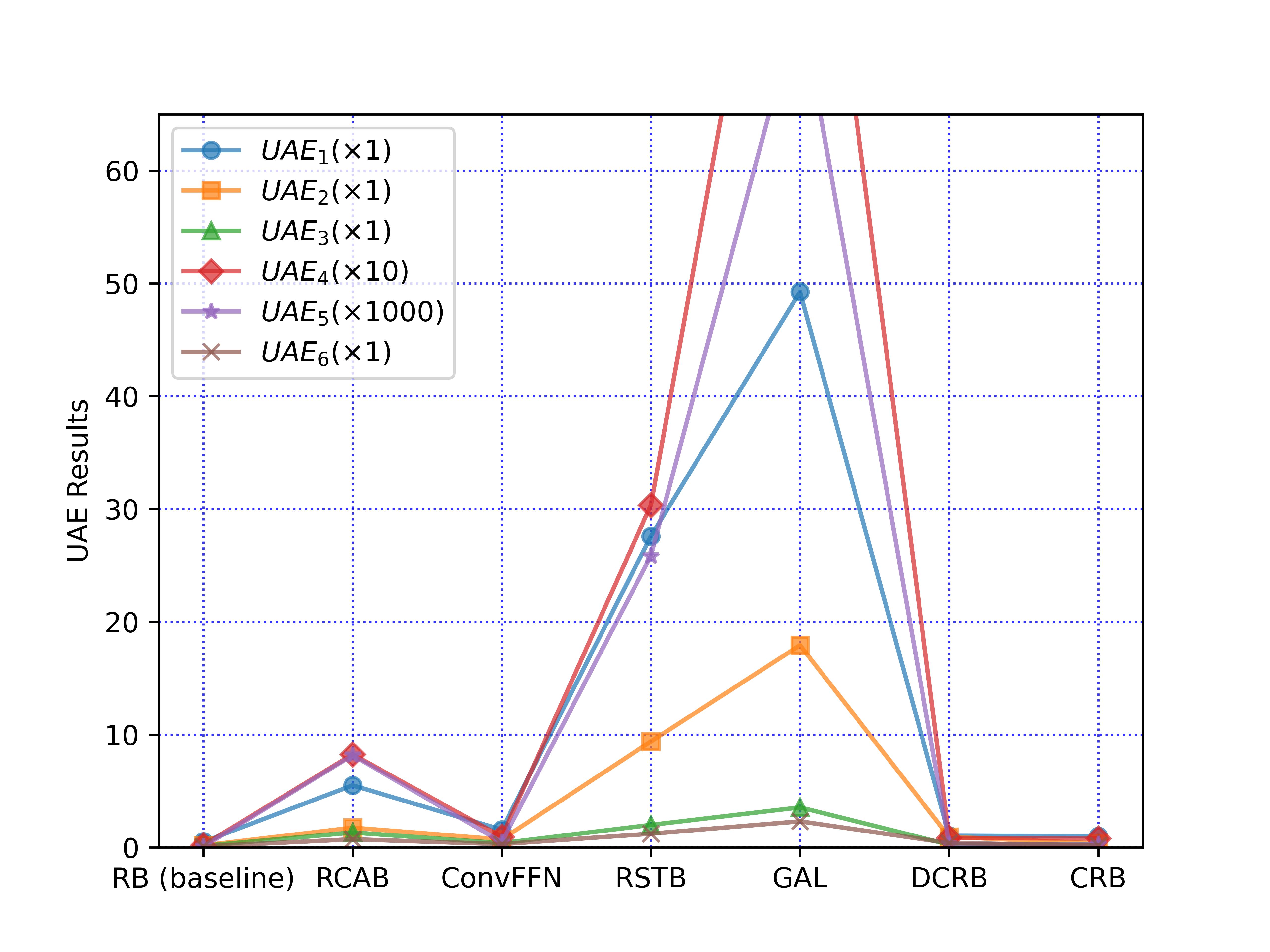}
		\caption{UAE  results on plug-and-play/non-plug-and-play modules.}
		\label{UAE visualization}
	\end{minipage}
	\begin{minipage}[t]{0.30\textwidth}
		\centering
		\includegraphics[width=\textwidth]{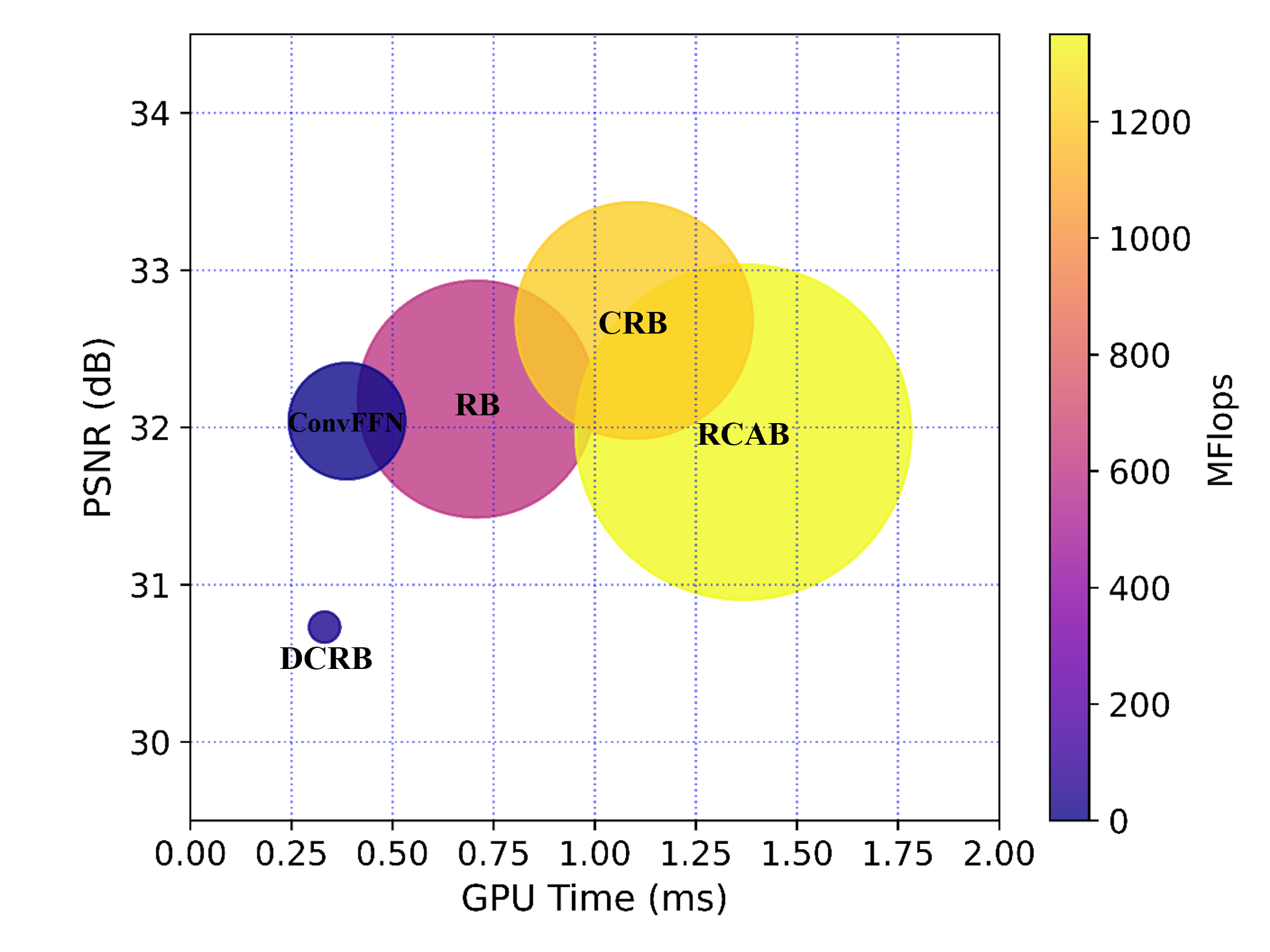}
		\caption{Computational overhead analysis on different plug-and-play modules. }
		\label{computational overhead}
	\end{minipage} 
	\begin{minipage}[t]{0.29\textwidth}
		\centering
		\includegraphics[width=\textwidth, trim=0cm 0cm 0cm 0.3cm, clip]{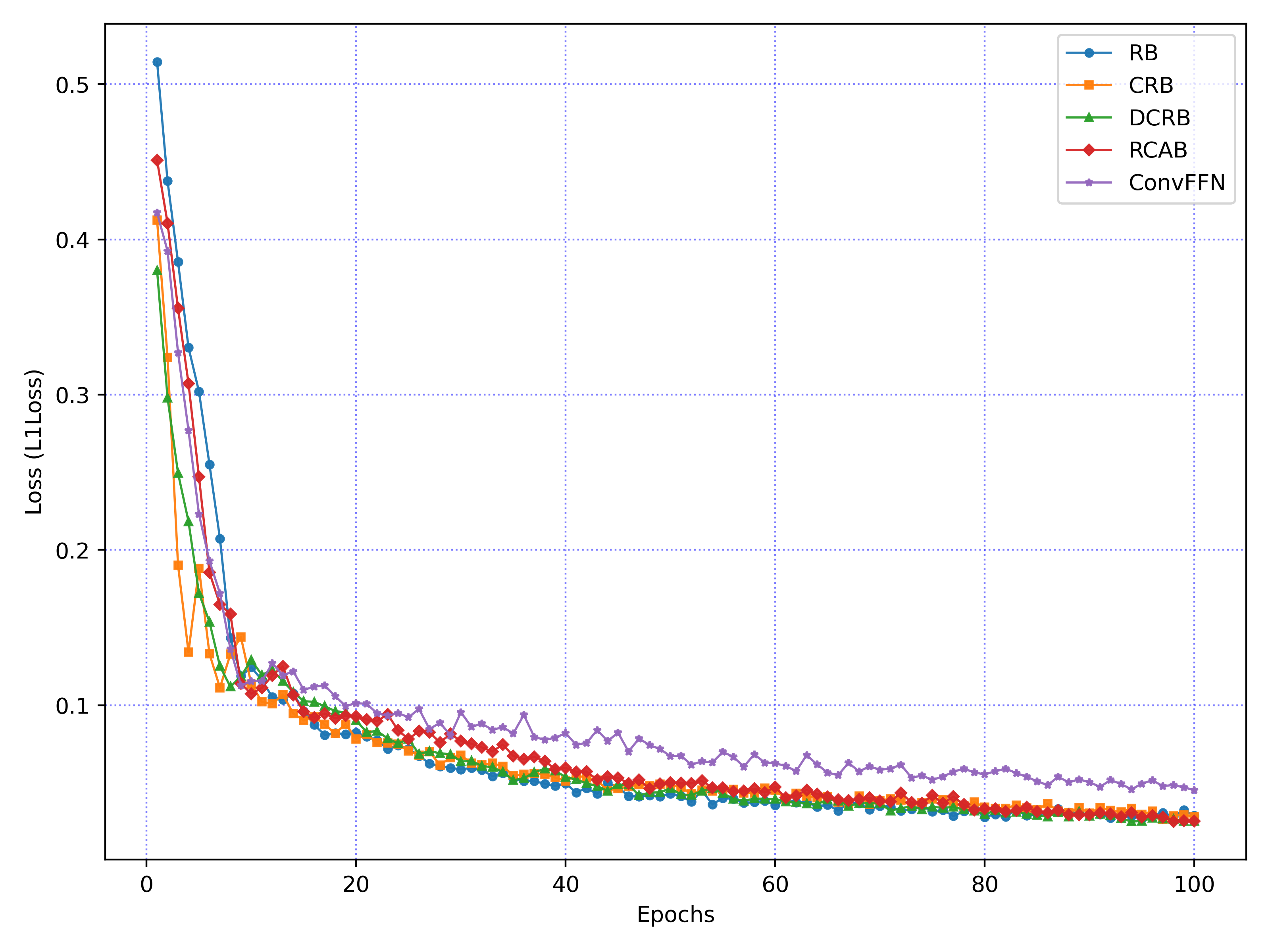}
		\caption{Convergence comparison between plug-and-play modules .}
		\label{convergence test}
	\end{minipage}
\end{figure*}

To validate Eq.\eqref{UAE abstract} and demonstrate that varying the form of UAE does not alter module universality rankings, we evaluate six distinct UAEs (Eq.\eqref{UAE forms}) on RB \cite{Resnet}, RCAB \cite{RCAN}, ConvFFN \cite{ViT}, RSTB \cite{SwinIR}, GAL \cite{IPG}, CRB and DCRB. The resulting universality scores are reported in Table \ref{Universality Assessment} and illustrated in Fig.\ref{UAE visualization}. 

From Fig.\ref{UAE visualization}, although the absolute values differ markedly due to the varying nature of each equation, the relative ordering remains invariant: $\phi(\text{GAL})>\phi(\text{RSTB})>>\phi(\text{RCAB})>\phi(\text{ConvFFN})>\phi(\text{DCRB})\approx\phi(\text{CRB})>\phi(\text{RB})$ where RB is adopted as the reference baseline. Hence, CRB and DCRB attain universality comparable to RB. Moreover, the intermediate magnitudes and consistent ranking of  $\phi_1$ and $\phi_2$ render analyses based on these two UAEs particularly transparent and interpretable.

Table \ref{Universality Assessment} demonstrates the core determinants of universality. Although RCAB \cite{RCAN} possesses larger $n$ and $l$ compared to RSTB \cite{SwinIR}, the elevated $k$ in RSTB deteriorates its transferability. This finding aligns with empirical evidence: RSTB integrates three nested submodules: MSA, MLP, and STL, whereas RCAB employs only the channel attention. Hence, effective migration of RSTB requires the concurrent adaptation of multiple components. Conversely, despite $n_{\mathrm{CRB}}\gg n_{\mathrm{DCRB}}$, DCRB remains more difficult to transfer owing to the inclusion of depth‐wise convolution, even though at a macroscopic level  $\phi(\text{DCRB})\approx\phi(\text{CRB})$.
\begin{equation}
	\begin{cases}
		\phi_1 = \frac{l \cdot e^{k+1}}{f} \cdot \lg\left(\frac{n}{100}\right) \\
		\phi_2 = \frac{l \cdot e^{k+1} \cdot \text{Sigmoid}\left(\frac{n}{100}\right)}{f} = \frac{l \cdot e^{k+1}}{f \cdot \left(1 + e^{-n/100}\right)} \\
		\phi_3 = \frac{l \cdot \lg\left(\frac{n}{100}\right)\cdot \text{Swish}(k+1)}{f}  = \frac{l \cdot \lg\left(\frac{n}{100}\right)}{f}  \frac{k+1}{1 + e^{-k-1}}\\
		\phi_4 = \frac{e^{k+1}\cdot \lg(\frac{n}{100})}{f}\cdot l^2 \\
		\phi_5 = \frac{(l^2+1)\cdot (5k^2+1)\cdot\sqrt{n+2}}{f} \\
		\phi_6 = \frac{\text{Softplus}(2k+1)\cdot\text{Softplus}(l)\cdot\text{Tanh}(\frac{n}{100})}{f}
	\end{cases}
	\label{UAE forms}
\end{equation}

In summary, an appropriately chosen UAE formulation serves to normalize score ranges, thereby facilitating direct comparisons across modules. Moreover, once Eq.\eqref{UAE abstract} is satisfied, alternative formulations of the UAE do not affect the module universality rankings.
\begin{figure*}[t]
	\centering
	\includegraphics[width=0.7\textwidth]{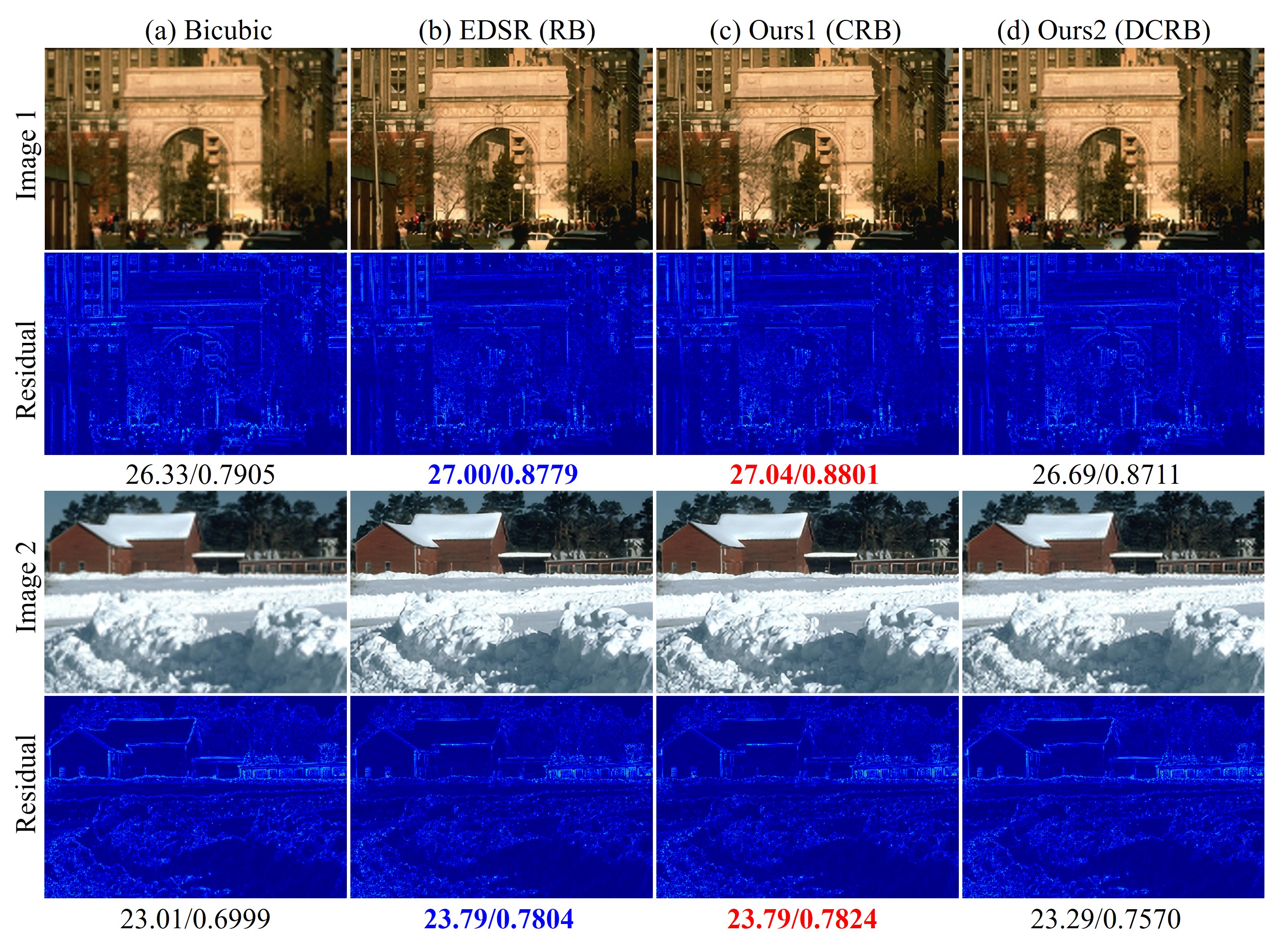}
	\caption{Visual comparison of conventional RB and optimized CRB/DCRB on 148089.png and 97033.png in B100 \cite{B100}.}
	\label{conventional visual}
\end{figure*}

\begin{figure*}[p]
	\centering
	\subfloat[729.png and 964.png in Potsdam \cite{potsdam}, where model* indicates that CRB is embedded, and the \textcolor{red}{red} highlights represent the best performance, while the \textcolor{blue}{blue} ones indicate the second-best.\label{remote sensing original}]{\includegraphics[width=0.9\textwidth]{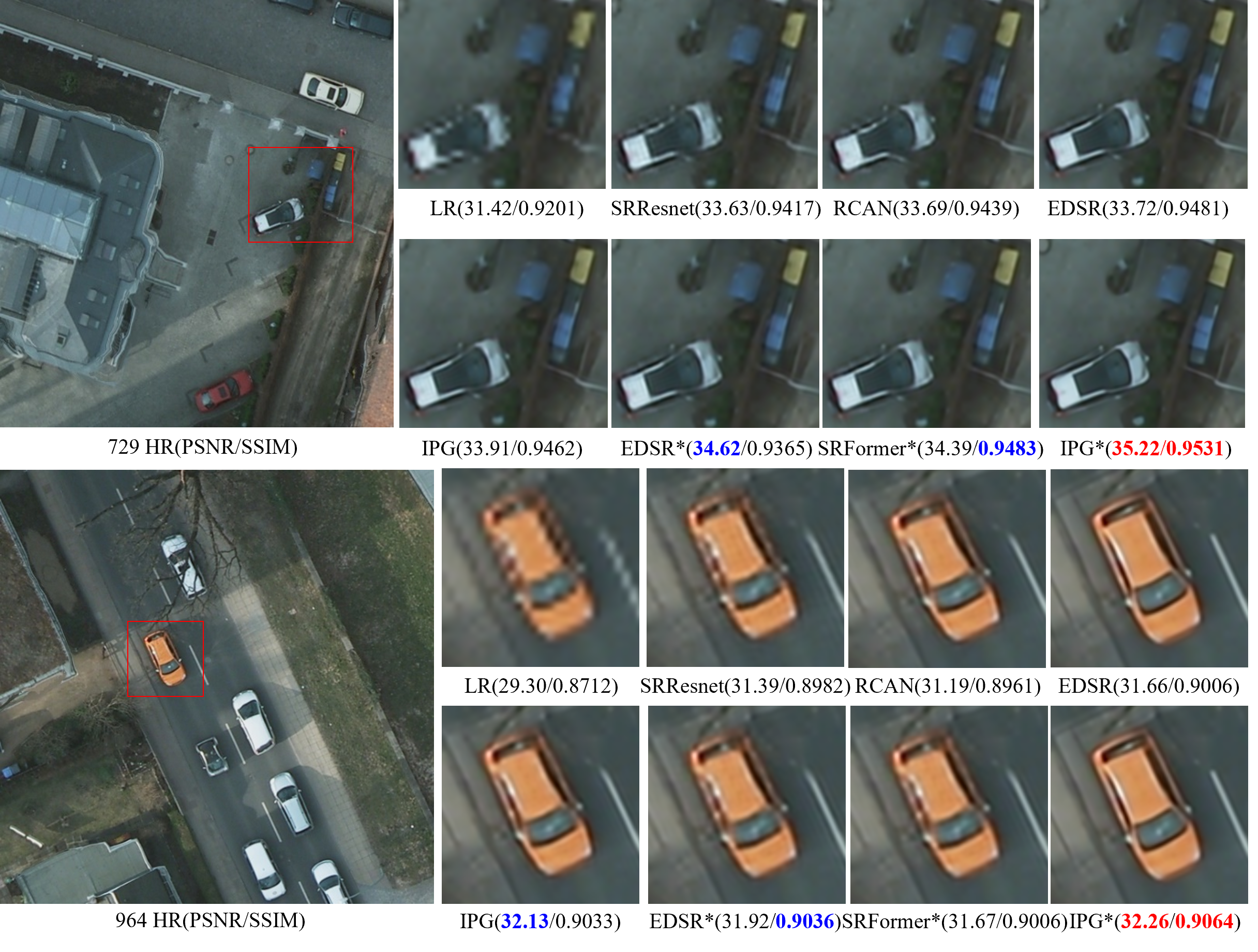}} \\
	\subfloat[Spatial maps of 898.png and 998.png in Potsdam  \cite{potsdam}, where the brighter regions indicate a greater perceptual loss between reconstructed images and its corresponding ground truth.\label{remote sensing spatial}]{\includegraphics[width=0.9\textwidth]{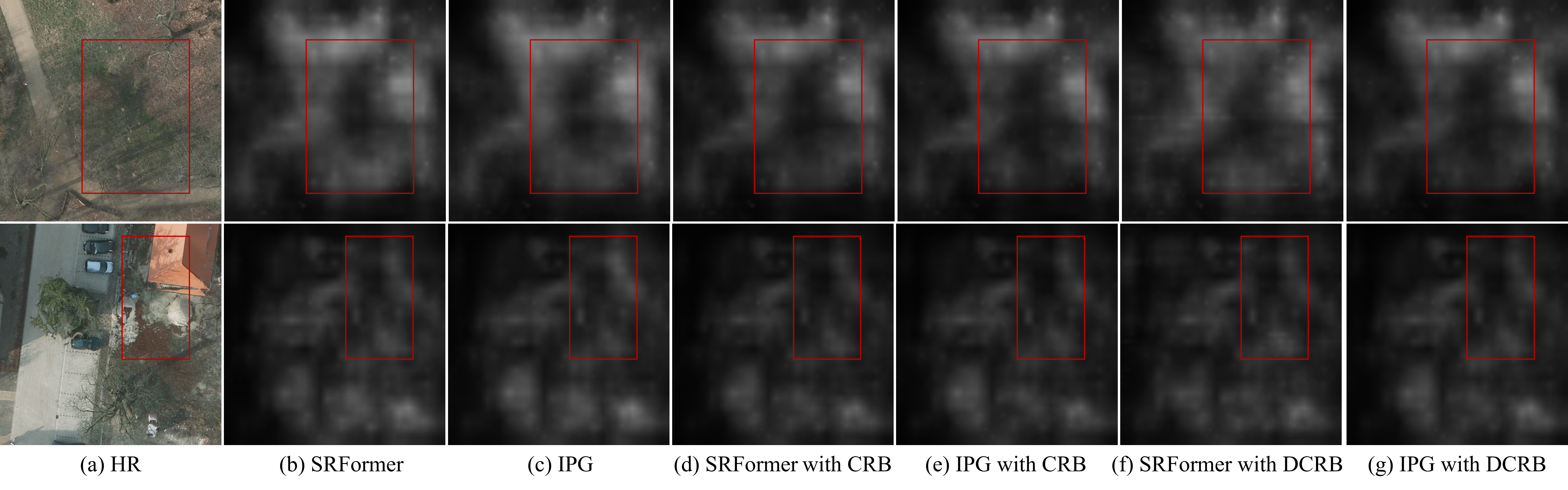}}
	\caption{Qualitative comparison of remote sensing images for $\times 4$ SISR.}
	\label{remote sensing visualization}
\end{figure*}

\begin{figure*}[p]
	\centering
	\includegraphics[width=\textwidth]{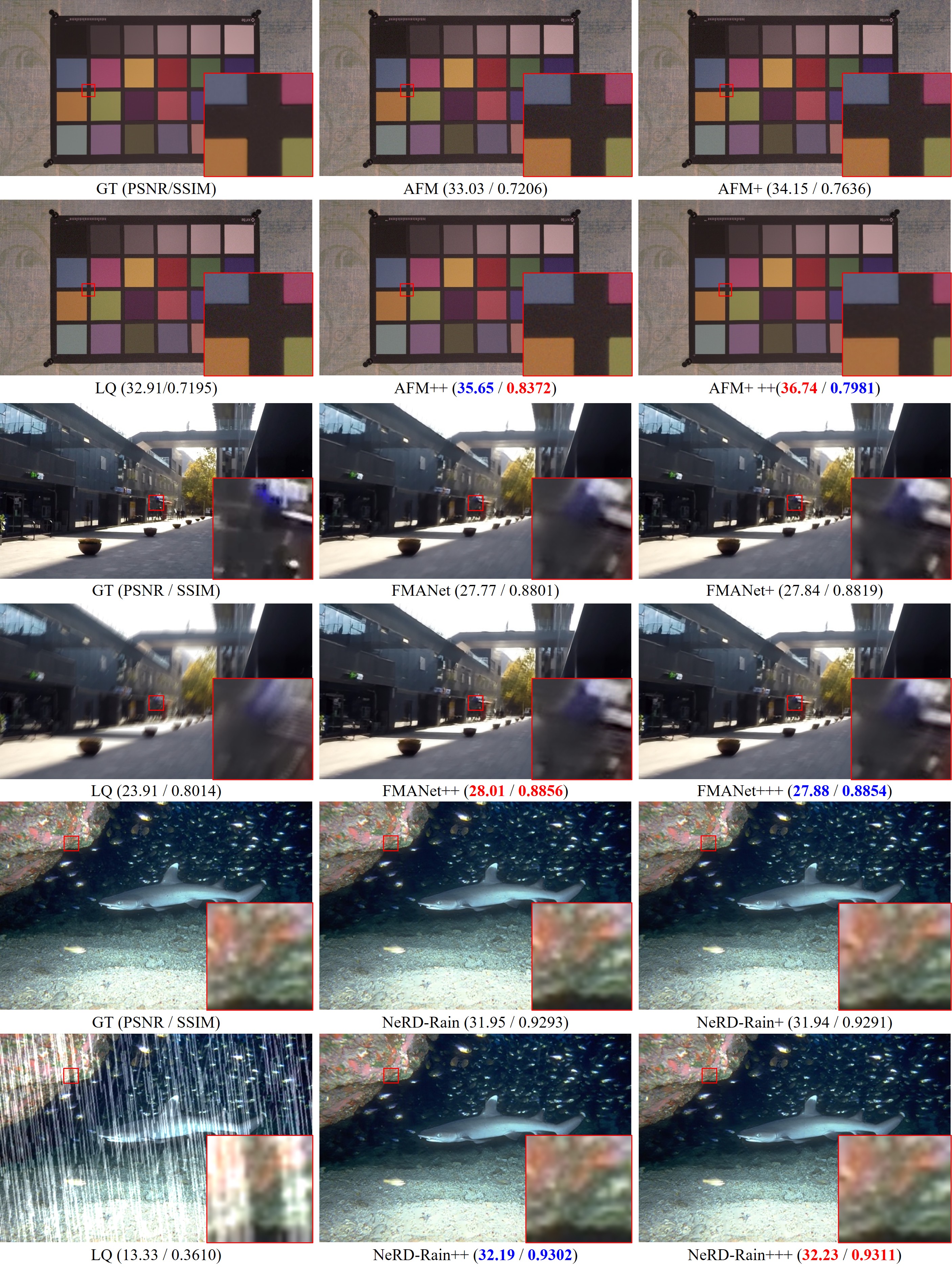}
	\caption{Visualization of the reconstructed results. ``Model+'' denotes insertion of the original plug-and-play module, ``Model++'' the UAE-optimized module, and ``Model+++'' the module further refined by both UAE and CRB. All embedded modules are in accordance with Table \ref{added}.}
	\label{other Low}
\end{figure*}
\subsection{Comparison Between UAE and Other Metrics}
\begin{table}[h]
	\centering
	\begin{tabular}{cccc}
		\toprule
		Metrics & Evaluation Object & Measurement Focus & Unit  \\
		\midrule
		Parameters & Model, Module & Storage requirements, Model capacity & K($10^3$), M($10^6$) \\
		FLOPs & Model, Module & Computational overhead, Inference speed & GFLOPs \\
		PSNR, SSIM & Model output & Reconstruction quality, Structural Similarity & dB, Dimensionless ([0,1]) \\
		Parameter efficiency & Model, Module & Performance per parameter & Metric per parameter \\
		GPU time & Model, Module & Inference time on hardware & ms \\
		\textbf{UAE (ours)} & Module & Structural flexibility, Transferability & Dimensionless ($[0,+\infty]$)  \\
		\bottomrule
	\end{tabular}
	\caption{Comparison between UAE and other existing metrics.}
	\label{Comparision}
\end{table}

\begin{figure}[h]
	\subfloat[Complete form ($\phi(l,k,n)$)\label{Spearman UAE}]{\includegraphics[width=0.5\textwidth]{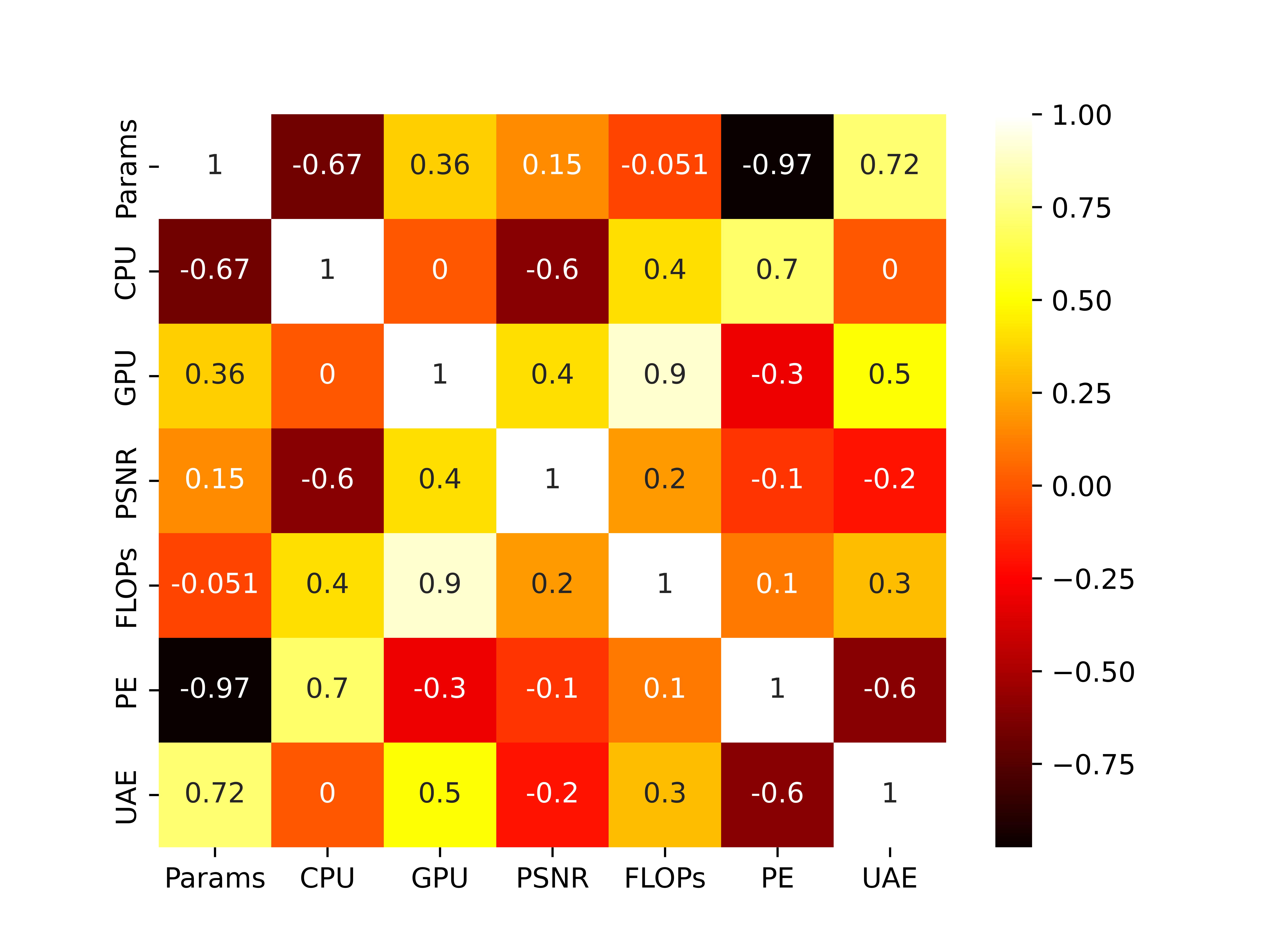}}
	\subfloat[$\alpha(l)$ only ($\phi(l)$)\label{Spearman alpha only}]{\includegraphics[width=0.5\textwidth]{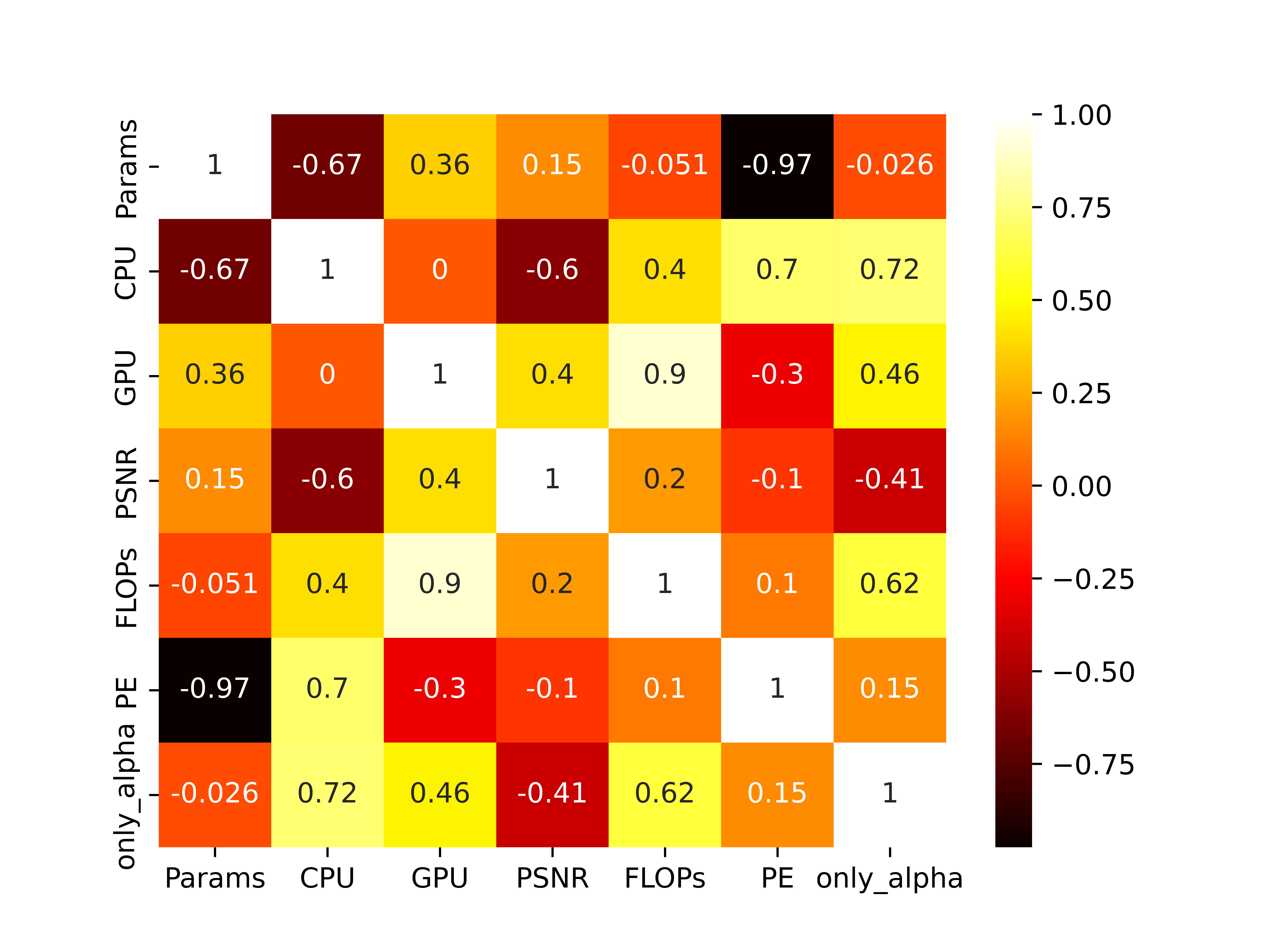}} \\
	\subfloat[$\beta(k)$ only ($\phi(k)$\label{Spearman beta only })]{\includegraphics[width=0.5\textwidth]{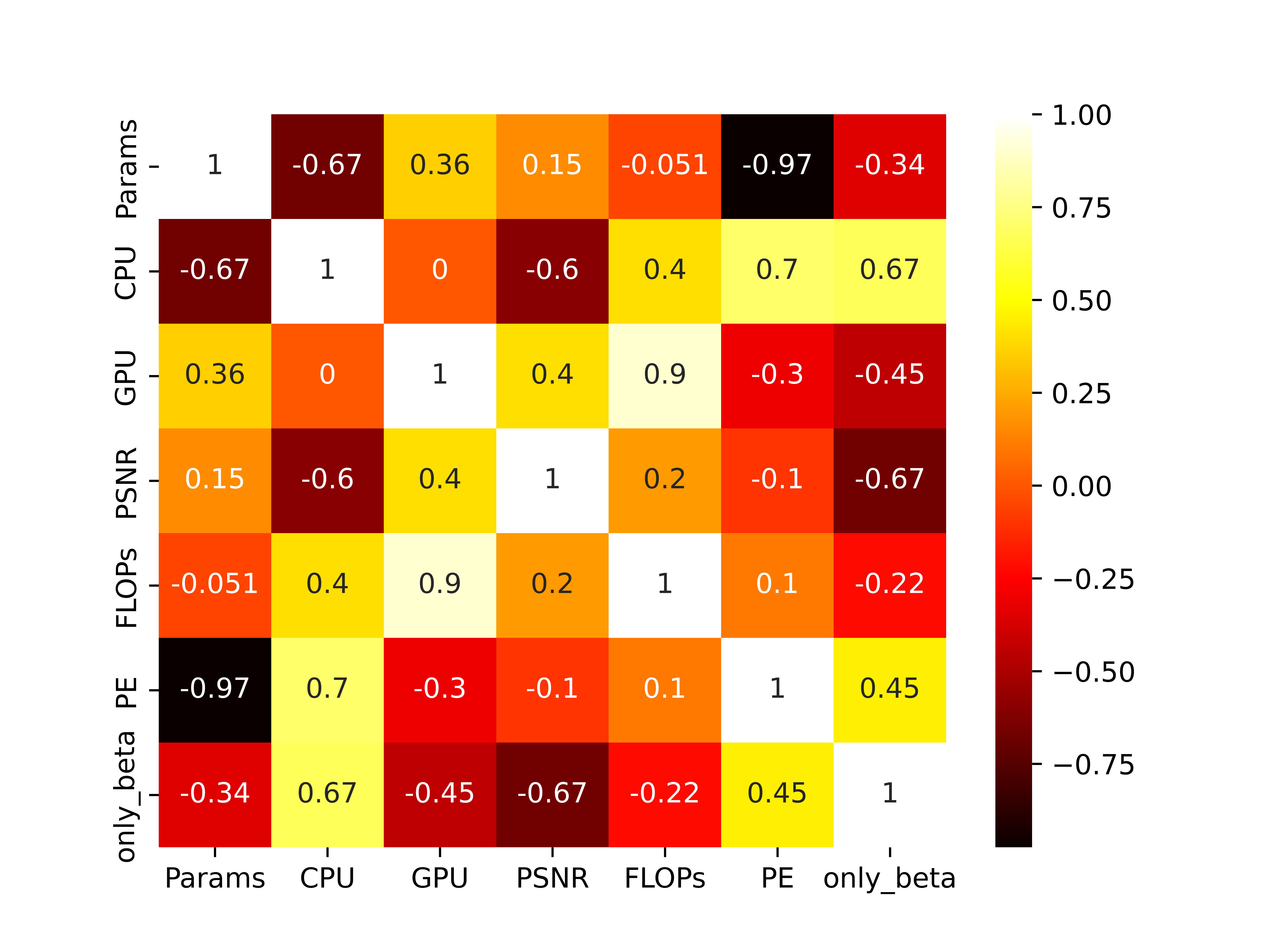}}
	\subfloat[$\theta(n)$ only ($\phi(n)$)\label{Spearman theta only }]{\includegraphics[width=0.5\textwidth]{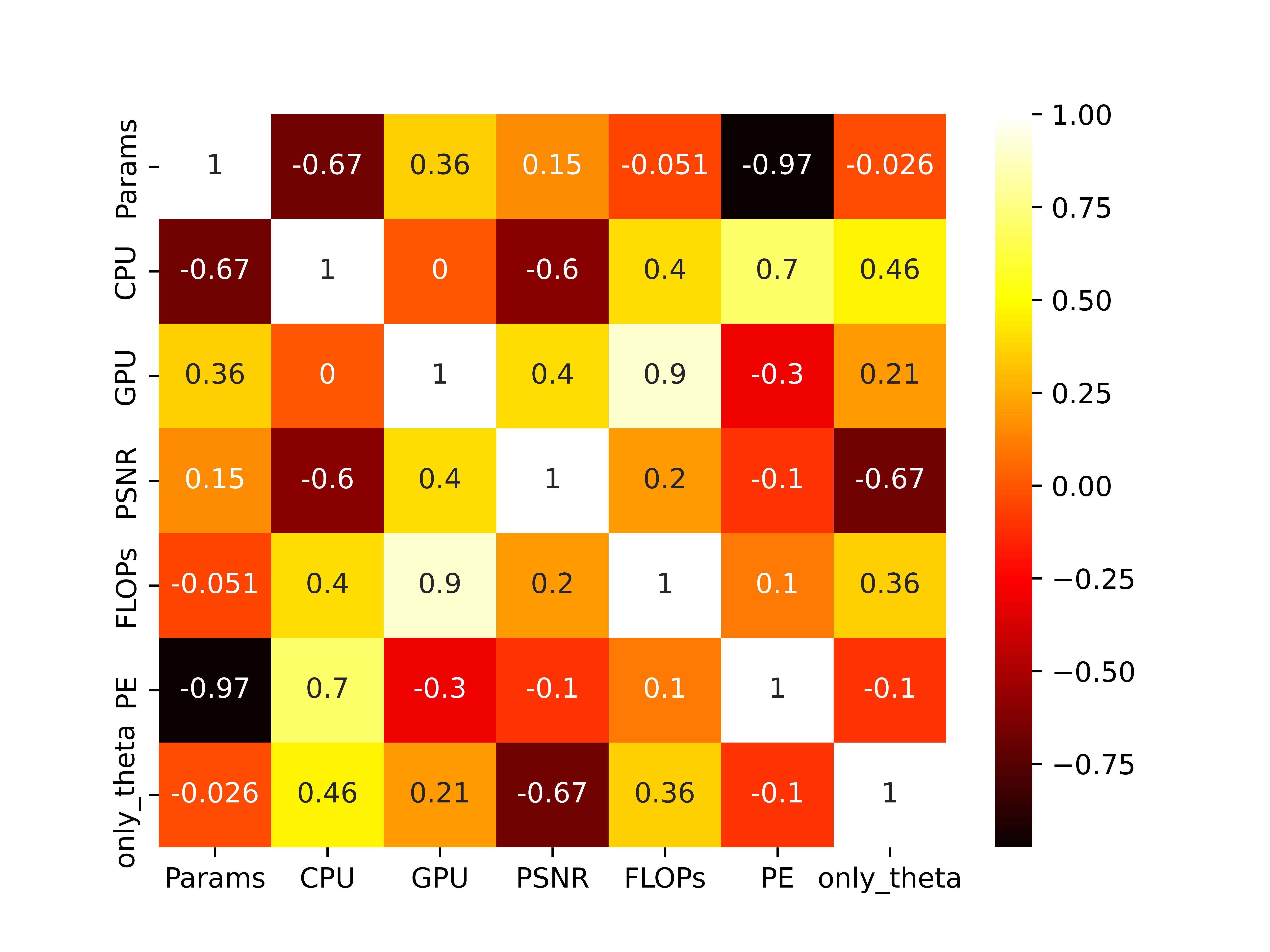}}
	\caption{Spearman Correlation Analysis on UAE. ``GPU'' represents modules' inference time on GPU, ``PE'' stands for parameter efficiency. Fig.\ref{Spearman alpha only}, \ref{Spearman beta only }, \ref{Spearman theta only } present the Spearman correlation analysis between each single component of UAE and other metrics, aiming to investigate which aspects of existing metrics are captured by $k,n,l$ in UAE. }
	\label{Spearman analysis}
\end{figure}

As shown in Table \ref{Comparision}, UAE differs from conventional metrics in several aspects. While existing metrics such as PSNR and GPU time primarily focus on performance evaluation at the model/module level, UAE is specifically designed to evaluate the structural flexibility and transferability of individual modules. This enables UAE to capture an essential yet previously unmeasured aspect of module design: its ability to migrate across different networks. 

Additionally, given that module transferability is affected by multiple factors, Fig.\ref{Spearman analysis} employs the Spearman Correlation Analysis to demonstrate that UAE is a composite metric integrating several metrics. Each single factor $\alpha, \beta, \theta$ exhibits moderate to strong correlation with certain existing metrics. For instance, $\alpha(l)$ correlates highly with CPU time  and FLOPs, while $\beta(k)$ has a moderate correlation with parameter efficiency.  $\beta$ exhibits a strong negative correlation with PSNR, because modules with very deep nesting levels fail to function independently and must operate within a complete model. However, UAE evaluates the performance of an isolated module to investigate its transferability. As a result, for modules with large $k$, the PSNR performance tends to decrease as the nesting depths increases, leading to a pronounced negative correlation.

The full UAE metric (Fig.\ref{Spearman UAE}) results demonstrate that UAE is not dominated by a single factor (e.g., 0.72 with parameters, 0.5 with GPU time), but rather offers a more balanced assessment that captures aspects not reflected by existing metrics alone. Therefore, UAE is a novel metric that reveals the combined influence of multiple existing metrics on transferability, an effect not captured by any individual metric.

\subsection{Computational Overhead Analysis}
As illustrated in Fig.\ref{computational overhead}, CRB achieves the highest PSNR, while DCRB delivers comparably strong fidelity with the lowest FLOPs, parameter count, and GPU time. Furthermore, CRB outperforms RB in PSNR under equivalent parameters (tested on Potsdam \cite{potsdam}). The convergence profiles in Fig.\ref{convergence test} indicate that all universal modules exhibit similar training patterns, with CRB attaining marginally faster convergence within the first 20 epochs, a behavior attributable to the accelerated gradient flow afforded by its additional residual branches.
\subsection{SISR for General Scene Images}
Using EDSR \cite{EDSR} as the backbone network, we conduct a combined analysis of module universality and performance, as shown in Table 
\ref{performance-plug-and-play}. Nearly all of the optimized modules feature better generalizability than baseline RB by slightly sacrificing their universality. Notably, RCAB--despite suffering the greatest universality degradation--fails to achieve the best performance. In contrast, CRB attains the largest generalization gains with minimal impact on universality, marking it as a positively universal module that could be integrated into diverse architectures without additional computational overhead. This finding demonstrates that CRB represents an exceptionally ideal module and indicates that universality and generalizability are not isolated. Although prior work has shown that higher generalization could be attained by sacrificing module transferability via an increased complexity and representational capacity, their precise interplay remains contingent on the specific architecture of plug‑and‑play modules.

Table \ref{multiple scale} summarizes CRB and DCRB performance across multiple SR scales. EDSR augmented by CRB consistently achieves the highest fidelity (PSNR/SSIM), whereas DCRB yields only minor performance degradations—-never exceeding a 5\% gap and reaching a minimum of 0.23\% under $\times 4$ SR on B100. Remarkably, DCRB comprises just 1.73\% of the parameter count of CRB and RB (Table \ref{Universality Assessment}), yet under certain conditions delivers approximately 99\% of the baseline performance.

As illustrated in Fig.\ref{conventional visual}, we present the $\times 3$ SR results on the B100 test set to visually compare CRB (Ours$_1$) and DCRB (Ours$_2$). Both variants achieve marked clarity enhancements over the LR, yielding similar qualitative reconstructions. The residual images reveal that CRB substantially reduces errors in critical details (e.g., roof eaves in 97033.png). In contrast, DCRB incurs increased residual magnitudes compared to the baseline, particularly within detailed texture areas where errors are more pronounced.

In summary, although DCRB offers superior transferability across model backbones, its positive impact on overall generalizability is limited. In contrast, CRB not only ensures robust plug‑and‑play universality but also substantially enhances performance across diverse datasets. Accordingly, CRB is best suited for applications with strict performance requirements, whereas DCRB is advantageous for lightweight or resource‑constrained deployments.

\subsection{SISR For Remote Sensing Images}
We employ the Potsdam dataset \cite{potsdam} for both training and testing. Benchmark models include SRCNN \cite{SRCNN}, SRResNet \cite{Resnet}, RCAN \cite{RCAN}, EDSR \cite{EDSR}, IPG \cite{IPG}, SRFormer \cite{SRFormer} as well as their enhanced variants integrating the CRB and DCRB modules.

Table \ref{remote sensing} reports the remote‐sensing SISR results. Models augmented with CRB consistently outperform their baseline counterparts, demonstrating CRB’s superior universality and its positive impact on model generalizability. In particular, ``IPG$+$CRB'' achieves the best performance. In contrast, owing to DCRB's depthwise separable convolution design, DCRB occasionally underperforms CRB.

Fig.\ref{remote sensing original} illustrates the SISR results of remote sensing imagery. Integrating CRB into the model substantially enhances the resolution of ground vehicle, producing more detailed textures and sharper edge contours. Fig.\ref{remote sensing spatial} displays the spatial maps of the SISR outputs, facilitating a perceptual comparison of model performance. Both CRB and DCRB effectively suppress bright regions, with the IPG variant augmented by CRB achieving the highest perceptual quality.

\subsection{UAE Optimization on Other Basic Modules and Low-Level Tasks}

\begin{table}[t]
	\centering
	\begin{tabular}{c|c|ccc}
		\toprule
		\multirow{6}{*}{ Deblurring} & \multirow{2}{*}{Models} & \multicolumn{3}{c}{GOPRO}  \\
		&   & PSNR & SSIM & LPIPS \\
		\cline{2-5}
		& FMANet (CVPR 2024) &27.57 & 0.8339&0.0783 \\
		& PPA (ICME 2024) + FMANet & \underline{27.71}& 0.8383& 0.0725\\
		& $\text{PPA}^{\text{*}}$ + FMANet & 27.69& \textbf{0.8398} & \underline{0.0719} \\
		& $\text{PPA}^{\text{**}}$ + FMANet & \textbf{27.72}& \underline{0.8387} & \textbf{0.0699}\\
		\midrule 
		\midrule
		\multirow{6}{*}{ Deraining} & \multirow{2}{*}{Models} & \multicolumn{3}{c}{Rain200H}  \\
		&   & PSNR & SSIM & LPIPS \\
		\cline{2-5}
		& NeRD-Rain (CVPR 2024) &30.13 &\underline{0.9145} &0.0430 \\
		& ELAB (ECCV 2022) + NeRD-Rain & 30.19& 0.9140&0.0428 \\
		& $\text{ELAB}^{\text{*}}$ + NeRD-Rain &\underline{30.23} &0.9143 &\underline{0.0426} \\
		& $\text{ELAB}^{\text{**}}$ + NeRD-Rain & \textbf{30.42}& \textbf{0.9161}&
		\textbf{0.0422 }\\
		\midrule 
		\midrule 
		\multirow{6}{*}{ Denoising} & \multirow{2}{*}{Models} & \multicolumn{3}{c}{SIDD-Medium}  \\
		&   & PSNR & SSIM & LPIPS \\
		\cline{2-5}
		& AFM (CVPR 2024) & 29.63&0.5843 &0.0093 \\
		& Agent-Attention (ECCV 2024) + AFM &30.61 &0.6180 &0.0091 \\
		& $\text{Agent-Attention}^{\text{*}}$ + AFM & \underline{31.76}&\underline{0.6920} &\underline{0.0089 }\\
		& $\text{Agent-Attention}^{\text{**}}$ + AFM &\textbf{32.56} &\textbf{0.7490 }&\textbf{0.0077} \\
		\midrule 
		\midrule 
		\multirow{6}{*}{Joint Denoising \& Low-Light-Enhancement} & \multirow{2}{*}{Models} & \multicolumn{3}{c}{VILNC}  \\
		&   & PSNR & SSIM & LPIPS \\
		\cline{2-5}
		& $\text{ZeroIG}^{\dag}$ (CVPR 2024) &10.17 &0.2952 &0.4708 \\
		& ARConv (CVPR 2025) + $\text{ZeroIG}^{\dag}$ & 12.13&0.3353 &\textbf{0.3809} \\
		& $\text{ARConv}^{\text{*}}$ + $\text{ZeroIG}^{\dag}$ &\underline{12.23} &\textbf{0.4346 }&0.3932 \\
		& $\text{ARConv}^{\text{**}}$ + $\text{ZeroIG}^{\dag}$ &\textbf{12.50} &\underline{0.3806} &\underline{0.3911} \\
		\bottomrule 
	\end{tabular}
	
	\caption{UAE optimization on different modules and performance evaluation of the optimized models on various low-level tasks. ``$\text{Module}^{\text{*}}$'' represents the module optimized by UAE, and ``$\text{Module}^{\text{**}}$'' represents the CRBs embedded on the basis of UAE optimization. Among the optimized variants, the structural parameters of ZeroIG are moderately reduced to adapt to limited computing resources (marked as $\text{ZeroIG}^{\dag}$), resulting in reduced model performance. However, model performance could still be compared by measuring the relative values of metrics between models.}
	\label{added}
\end{table}

Firstly, we briefly summarize the models and corresponding UAE optimization methods for the four tasks. Since our UAE optimization stays only within the module, our methods do not affect the inherent plug-and-play characteristics of the modules. We only select appropriate places to embed our optimized components into the framework without changing models' core architectures.

\textbf{FMANet} (Flow-Guided Dynamic Filtering and Iterative Feature Refinement with Multi-Attention Network) \cite{FMANet} comprises a multi-layer convolutional feature extractor followed by point-wise mappings, emphasizing per-pixel enhancement via stacked convolution blocks. \textbf{PPA} (Parallelized Patch-aware Attention) \cite{PPA} is a multi-branch convolutional block fused with local/global attention and spatial attention, used for multi-scale context aggregation. During our UAE optimization process, we augment PPA with channel expansion, parallel depth-wise branches and branch fusion followed by spatial attention to increase the representational capacity. CRB is further inserted after $\text{PPA}^{\text{*}}$ output as a post-refinement stage.

\textbf{NeRD-Rain} (Neural Representations for Image Deraining) \cite{NeRD} is a multi-level architecture built from window processing, using local filters and stacked convolutions for deraining. \textbf{ELAB} (Efficient Long-rang Attention Block) \cite{ELAB} is a block combining local filtering and gated multi-window self-attention (GMSA), excelling at local detail refinement. When designing $\text{ELAB}^{\text{*}}$, we add a lightweight local depth-wise branch and SE-style channel reweighting to better use input features, while the GMSA structure is fully preserved. Further, CRBs are applied after $\text{ELAB}^{\text{*}}$ as a bidirectional residual refinement for detail correction. 

\textbf{AFM} (Adversarial Frequency Mixup Framework) \cite{AFM} is a lightweight DnCNN-style backbone, consisting of multiple convolution layers that perform per-pixel noise restoration. \textbf{Agent-Attention} \cite{Agent-Attention} derives adaptive pooled agent tokens and injects agent information into features. We optimize Agent-Attention by increasing heads and agent numbers, and we further embed CRBs inside the agent wrapper, where the attention output is projected and then passed through CRB before normalization and residual additions. 

\textbf{ZeroIG} (Zero-Shot Illumination Guided Framework) \cite{ZeroIG} is a decomposition-based architecture with Enhancer and Denoising modules arranged in a cascaded fashion, which could perform denoising and low-light-enhancement simultaneously. \textbf{ARConv} (Adaptive Rectangular Convolution) \cite{ARConv} is an adaptive resampling convolution module which fixes the drawbacks of convolution operations within a confined square window. We further wrap the ARConv with channel expansion and multiple parallel branches to better leverage the input features. CRBs are embedded in ZeroIG's high-frequency branch (H3). To be specific, after the denoising step, CRBs are applied for detail refinement.

Table \ref{added} demonstrates that models embedded with the plug-and-play modules exhibit better performance than the baselines, and after our UAE optimization, the performance has generally been further improved. Additionally, models embedded with the modules that are optimized by both UAE and CRB have achieved new SOTAs on nearly all the datasets.

From a qualitative perspective, as shown in Fig. \ref{other Low}, images restored by the optimized models also have better visual quality than the baselines, and most of the details in LQ (Low Quality) images have been recovered. Furthermore, as shown in the deraining part of Fig. \ref{other Low},  although the performance of NeRD-Rain+ is slightly degraded after directly embedding the plug-and-play module into NeRD-Rain, after UAE optimization and CRB enhancement, however, the model performance still surpasses the baseline, providing direct evidence for the effectiveness of our UAE optimization strategy.

Therefore, our optimization strategy is not only applicable to SR tasks, but also other low-level tasks such as denoising, deblurring, and deraining.

\subsection{Ablation Studies}
\subsubsection{Sensitivity Quantification of UAE Parameters}
\begin{table*}[h]
	\centering
	\begin{tabular}{cccccccccc}
		\toprule
		$\alpha(l)$ & $\beta(k)$ & $\theta(n)$ & RB & RCAB & ConvFFN & RSTB & GAL & DCRB & CRB \\ 
		\midrule
		\checkmark & & & 0.063 & 0.234 &  0.094 & 0.172 & 0.328 & 0.125 & 0.125 \\
		& \checkmark &  & 0.011 & 0.028 & 0.028 & 0.061 & 0.061 & 0.028 & 0.030 \\
		& & \checkmark & 0.045 & 0.050 & 0.035 &0.046 & 0.043 & 0.017 & 0.045 \\
		& \checkmark & \checkmark & 0.033&0.087 & 0.062 & 0.180 & 0.169 & 0.030 & 0.033  \\
		\checkmark & & \checkmark & 0.179 & 0.743 &  0.211 & 0.505 & 0.902 & 0.138 & 0.359 \\ 
		\checkmark & \checkmark & & 0.046 & 0.413 & 0.165 & 0.675 & 1.289 & 0.220 & 0.091 \\
		\checkmark & \checkmark & \checkmark & 0.131 & 1.309 & 0.372 & 1.984 & 3.543 & 0.244 & 0.262 \\
		\bottomrule
	\end{tabular}
	\caption{Variable Ablation of UAE, where $\phi$\textsubscript{3} is specified as the UAE form and the module parameters refer to Table \ref{Universality Assessment}.}
	\label{Variable Ablation}
\end{table*}

A logarithmic derivative-based approach and elasticity coefficients are proposed to quantify the variable sensitivity. For a general UAE form where $\phi = \Pi_i f_i(x_i)$, the logarithmic differentiation goes as Eq.\eqref{logarithmic derivativeness}

\begin{equation}
	d\ln{\phi}=\sum_{i}\frac{\partial \ln{f_i}}{\partial \ln{x_i}}d\ln{x_i}
	\label{logarithmic derivativeness}
\end{equation}

The sensitivity of $x_i$ is defined as the absolute elasticity and its normalized form, as shown in Eq.\eqref{Elasticity}. $S_i$ represents the percentage change in $\phi$ caused by a 1\% change in $x_i$. $x_i$ and $f_i$ respectively stand for the variables and functions in UAE. For instance, $\{<x_{\alpha},f_{\alpha}>\}=\{<l,\alpha(l)>\}$. 
\begin{equation}
	S_i=\left|\frac{\partial \ln{\phi_i}}{\partial \ln{x_i}}\right|=\left|\frac{x_i}{f_i}\cdot\frac{\partial f_i}{\partial x_i}\right|,\hat{S_i} = \frac{S_i}{S_{\alpha}+S_{\beta}+S_{\theta}},i\in\{\alpha, \beta, \theta\}
	\label{Elasticity}
\end{equation}

Additionally, we employ the Sharply value to measure the contribution of each variable, which differs from  the sensitivity. Sharply value is a concept from cooperative game theory that allocates the total contribution of each participant by averaging their marginal contributions across all combinations. Given that UAE features three variables, the Sharply equations are defined as Eq.\eqref{Sharply}.
\begin{equation}
	\begin{cases}
		C_{\alpha} = \frac{v(\alpha)-v(\varnothing)+v(\alpha, \beta)-v(\beta)+v(\alpha,\theta)-v(\theta)}{3} \\
		C_{\beta} = \frac{v(\beta)-v(\varnothing)+v(\alpha, \beta)-v(\alpha)+v(\beta,\theta)-v(\theta)}{3} \\
		C_{\theta} = \frac{v(\theta)-v(\varnothing)+v(\theta, \beta)-v(\beta)+v(\alpha,\theta)-v(\alpha)}{3} 	
	\end{cases}
	\label{Sharply}
\end{equation}

\begin{figure*}[h] 
	\centering
	\subfloat[Absolute values of UAE ablation.\label{Variable Ablation 1}]{	\includegraphics[width=0.33\textwidth]{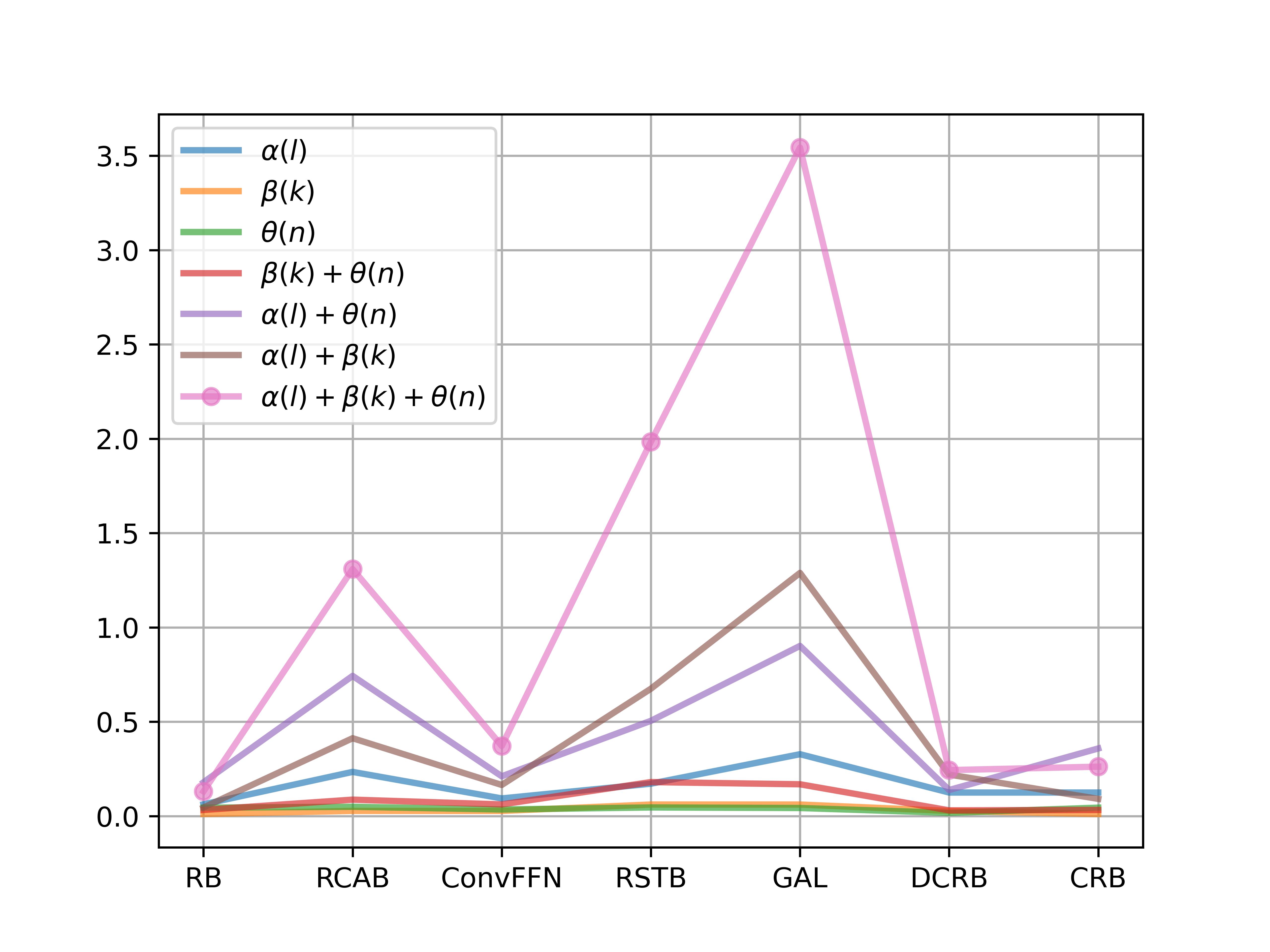}}
	\subfloat[Normalized sensitivity of $k,n,l$ for modules' UAE results.\label{Variable Ablation 2}]{	\includegraphics[width=0.33\textwidth]{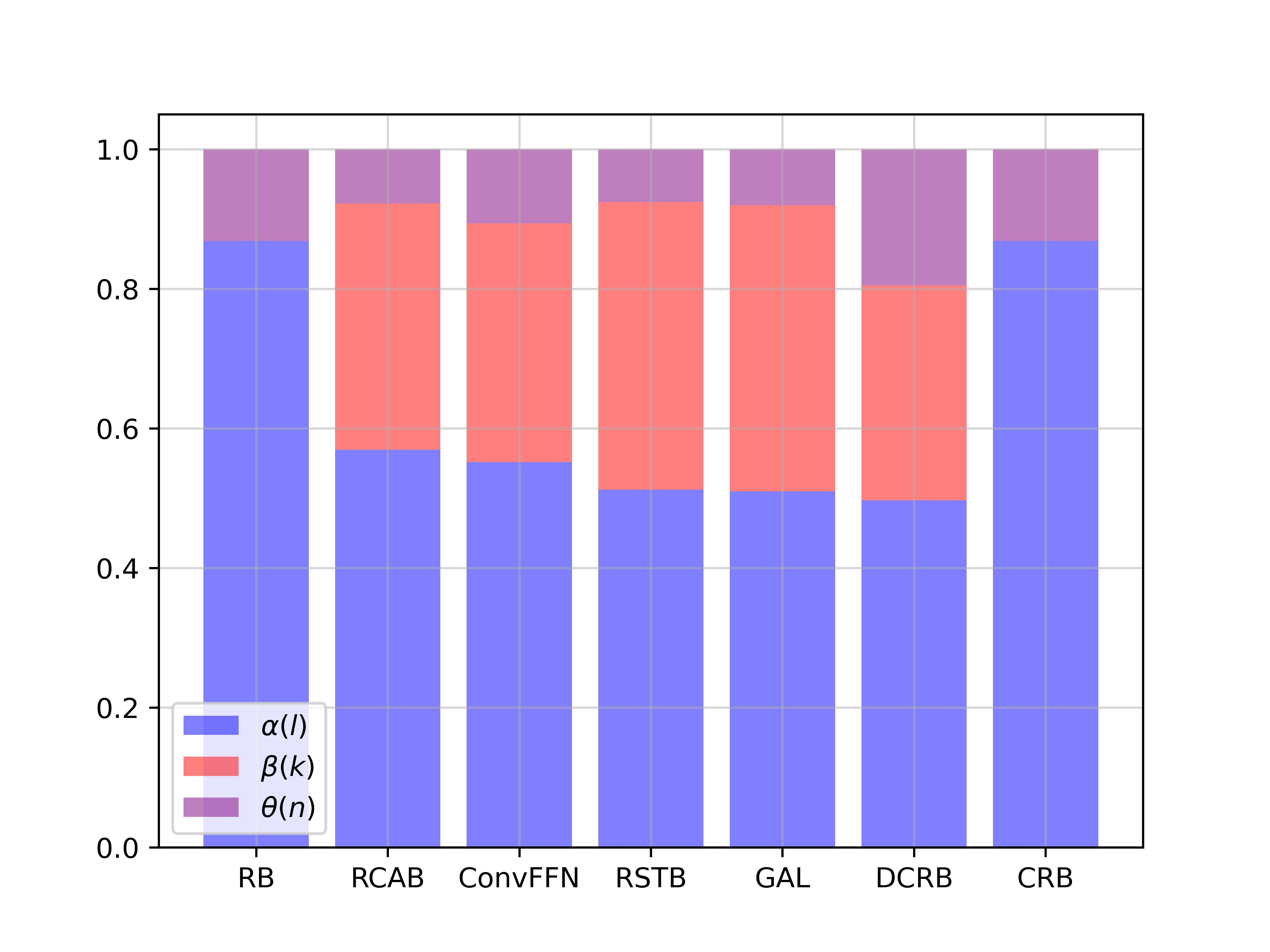}}
	\subfloat[Normalized contribution of $k,n,l$ for modules' UAE results. \label{Variable Abaltion 3}]{\includegraphics[width=0.33\textwidth]{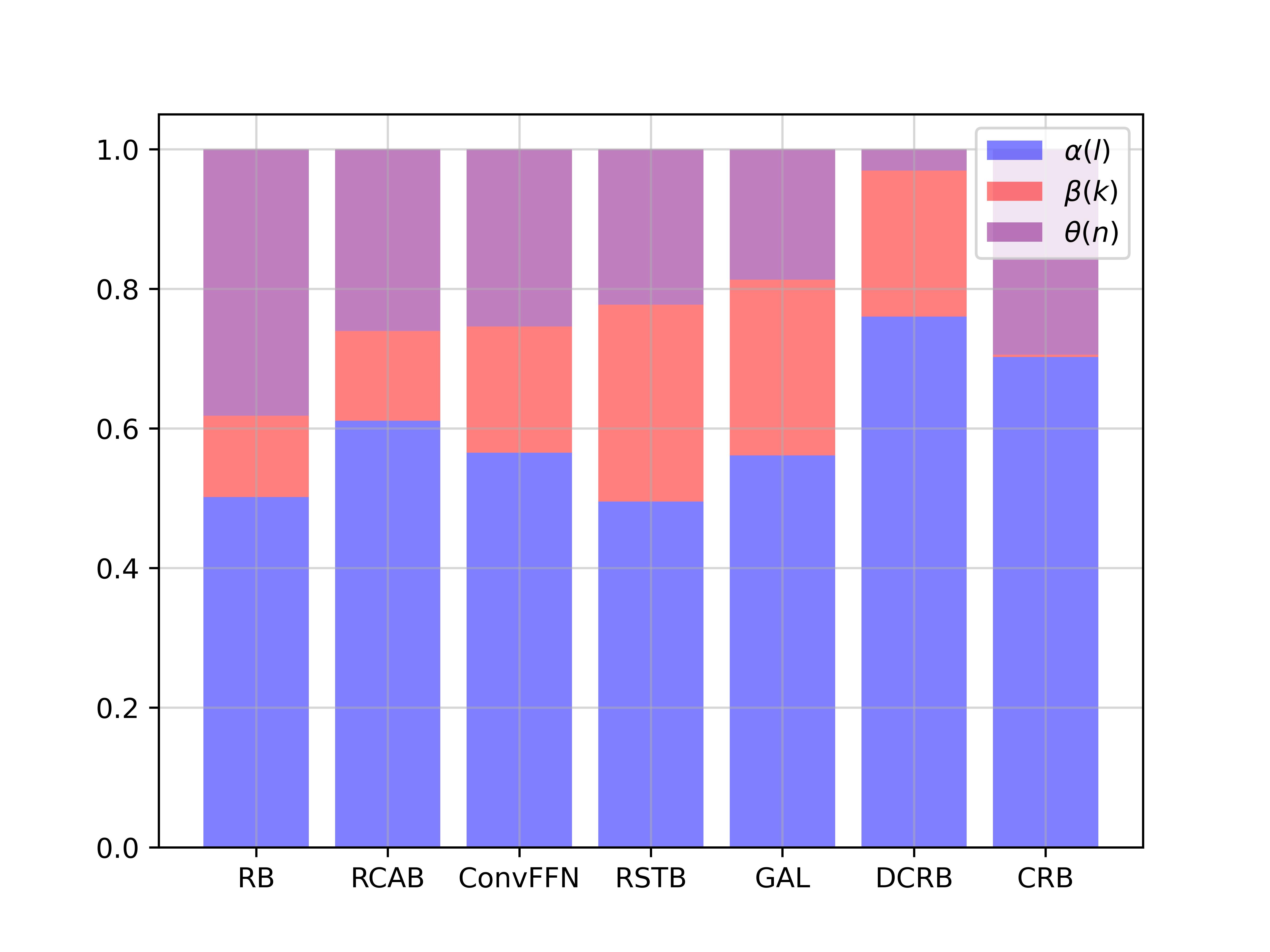}}
	\caption{Variable Ablation of UAE. Fig.\ref{Variable Ablation 1} demonstrates the effects of $x,<x,y>,<x,y,z> (x,y,z\in\{k,n,l\})$ on UAE calculation, and Fig.\ref{Variable Ablation 2} depicts the variable sensitivity via elasticity coefficients. Fig.\ref{Variable Abaltion 3} describes the absolute variable contributions on UAE.}
	\label{Variable Ablation Total}
\end{figure*}

Fig.\ref{Variable Ablation 2} quantitatively demonstrates that increases in $k$ induce nonlinear growth in module sensitivity to nesting depths, with the $\beta_{\text{Sens.}}$ escalating from 0.02 (k=0, RB) to 0.41 (k=3, GAL), a 20$\times$ amplification that satisfies the constraint $\beta''(k)>0$. Parameter $\theta_{\text{Sens.}}$ exhibits logarithmic saturation for large modules, evidenced by near-identical sensitivities for RSTB ($n=86,764$, $\theta_{\text{Sens.}}=0.076$) and GAL ($n=56,132$, $\theta_{\text{Sens.}}=0.081$). While Fig.\ref{Variable Abaltion 3} indicates that $l$ dominates absolute UAE contributions, sensitivity analysis still reveals $k$'s critical role as the primary leverage point: $[\beta_{\text{Sens.}}(k=3,GAL)-\beta_{\text{Sens.}}(k=0,RB)]>[\alpha_{\text{Sens.}}(l=21, GAL)-\alpha_{\text{Sens.}}(l=4, RB)]>[\theta_{\text{Sens.}}(n=56132, GAL)-\theta_{\text{Sens.}}(n=73856,RB)]$. Therefore, $k$-reduction is the foremost transferability optimization strategy for nested architectures, followed by $l$- and $n$-adjustments.

\subsubsection{Nature of CRB Gradient Explosion and Experimental Validation on the $\mathbf{\varepsilon}$ Constraints}
\begin{table*}[h]
	\centering
	\begin{tabular}{ccccccc}
		\toprule
		$l$ & $\varepsilon$ & $\varepsilon\leqslant (l/4 + \delta_{\text{safe}})$ &  $\mathcal{L}_1^{\text{min}}$ &Set5 (PSNR/SSIM) & Set14 (PSNR/SSIM)  & B100 (PSNR/SSIM) \\
		\midrule
		\multirow{3}{*}{8} & 1 &\Checkmark
		&7.31&27.39/0.8610&24.17/0.7470&23.96/0.7129  \\
		& 2 &\Checkmark&7.47&27.36/0.8607&24.15/0.7469&23.95/0.7128\\
		& 3 &\ding{55}&15.39&21.46/0.5703&20.58/0.5203&21.93/0.5220 \\
		\midrule
		\multirow{3}{*}{12} & 2 &\Checkmark&7.57&27.53/0.8643&24.25/0.7503&23.94/0.7126 \\
		& 3 &\Checkmark&7.50&27.32/0.8601&24.13/0.7468&23.98/0.7143 \\
		& 4 &\ding{55}&27.06&17.50/0.3236&17.46/0.3062&19.18/0.3427 \\
		\midrule
		\multirow{3}{*}{16} & 3 &\Checkmark&8.00&27.41/0.8622&24.18/0.7486&23.95/0.7133 \\
		& 4 &\Checkmark&7.77&27.46/0.8629&24.21/0.7492&23.96/0.7135 \\
		& 5 &\ding{55}&101.92&10.53/0.0320&11.22/0.0349&13.00/0.0629 \\
		\bottomrule
	\end{tabular}
	\caption{Ablation study of $\varepsilon$. Models are trained for 400 epochs on DIV2K dataset, EDSR is used as the backbone. $\delta_{\text{safe}}$ is set to zero during the experiments. $\mathcal{L}_1^{\text{min}}$ represents the minimum L1 loss during training.}
	\label{ablation epsilon}
\end{table*}

As shown in Table \ref{ablation epsilon}, when $\varepsilon$ exceeds the boundary $l/4+\delta_{\text{safe}}$, module performance significantly deteriorates, and when $\varepsilon\leqslant l/4+\delta_{\text{safe}}$, module performance remains remarkably similar. Therefore, excessive residual connections do not further enhance model performance.
\begin{figure*}[h]
	\centering
	\subfloat[$l=8,\varepsilon=2 (\varepsilon\leqslant l/4+\delta_{\text{safe}})$\label{CRB_8l_2epsilon} ]{\includegraphics[width=0.33\textwidth]{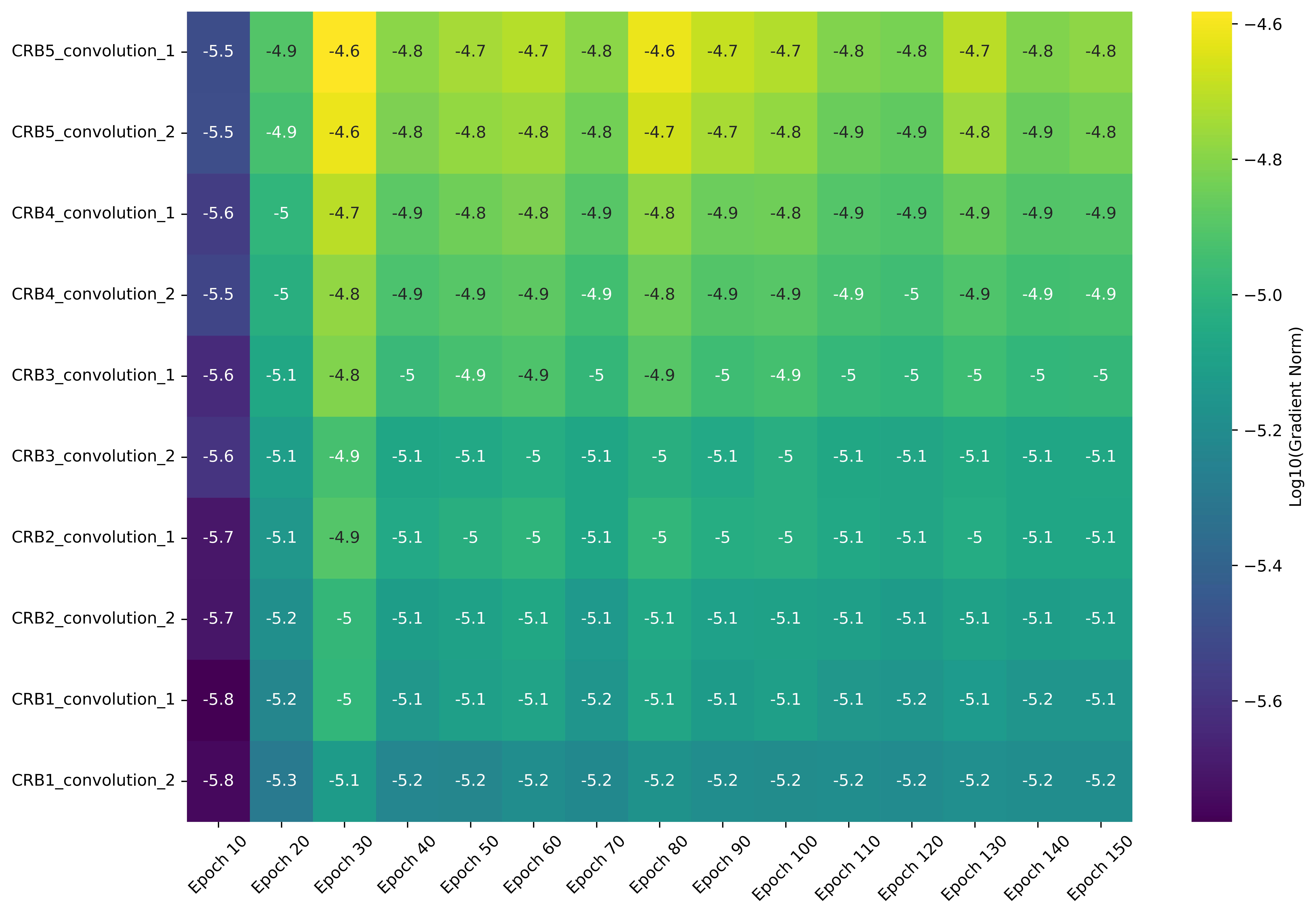}}
	\subfloat[$l=12,\varepsilon=3 (\varepsilon\leqslant l/4+\delta_{\text{safe}})$\label{CRB_12l_3epsilon}]{\includegraphics[width=0.33\textwidth]{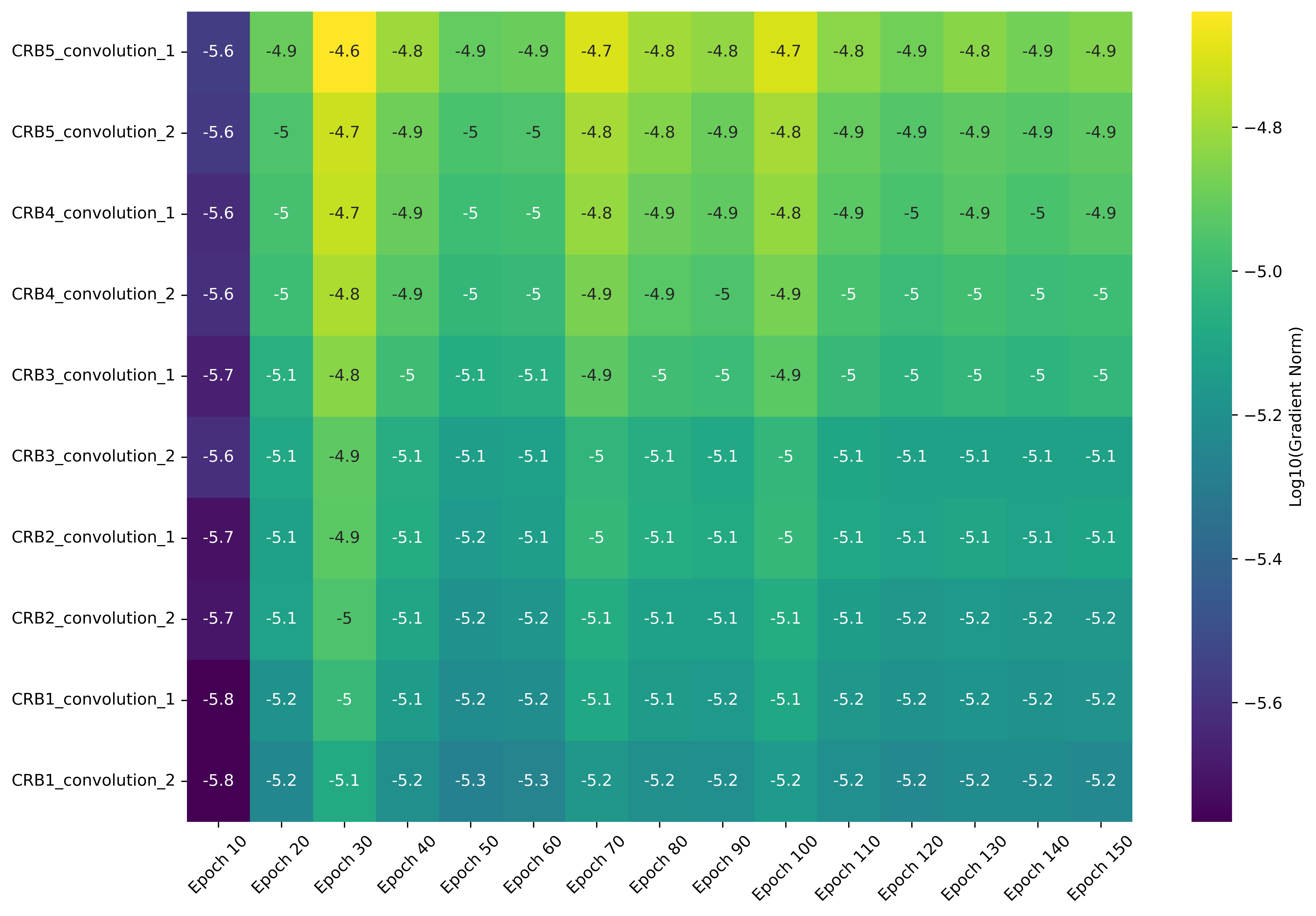}}
	\subfloat[$l=16,\varepsilon=4 (\varepsilon\leqslant l/4+\delta_{\text{safe}})$\label{CRB_16l_4epsilon}]{\includegraphics[width=0.33\textwidth]{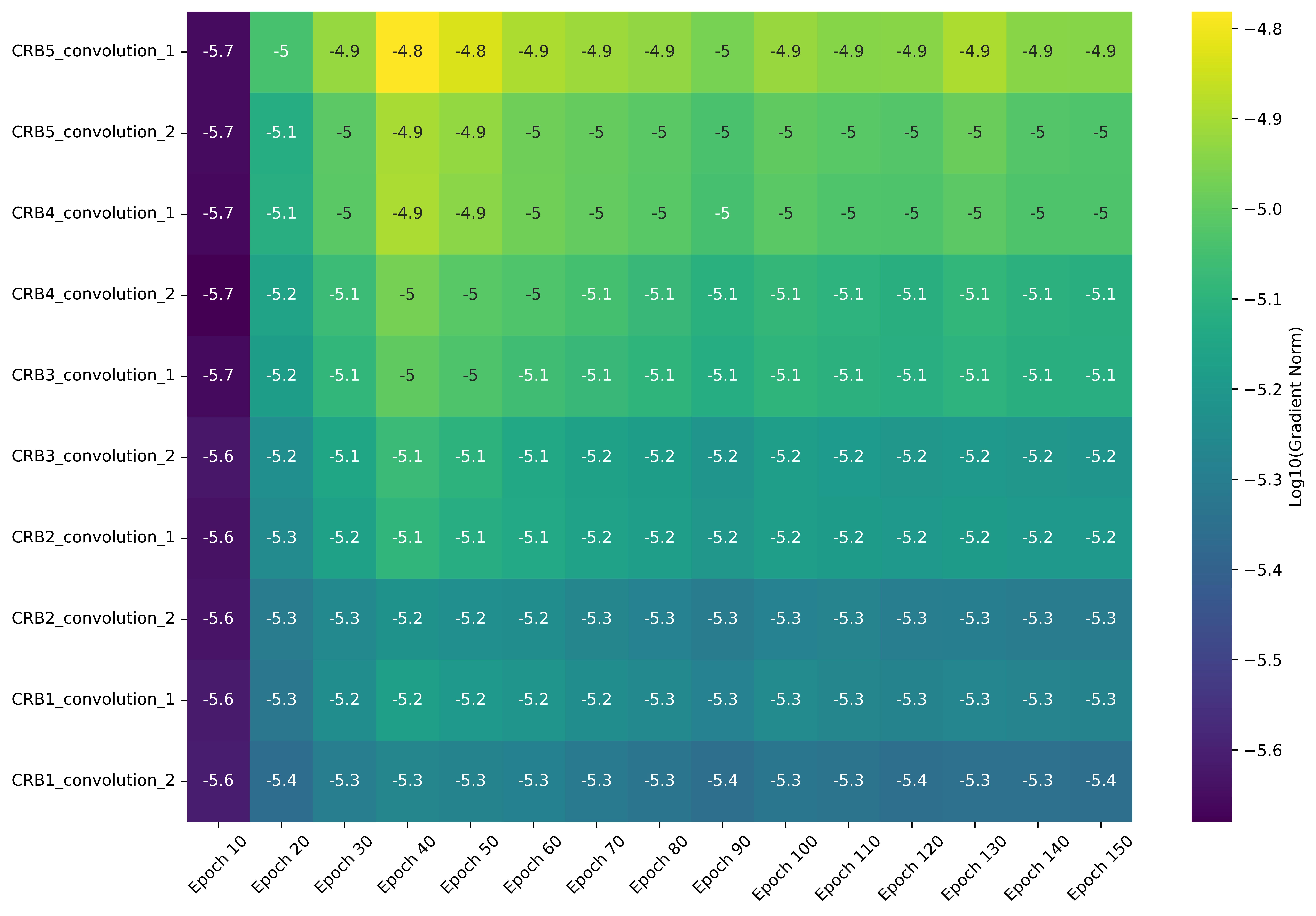}}\\
	\subfloat[$l=8,\varepsilon=3 (\varepsilon> l/4+\delta_{\text{safe}})$\label{explosion1}]{\includegraphics[width=0.33\textwidth]{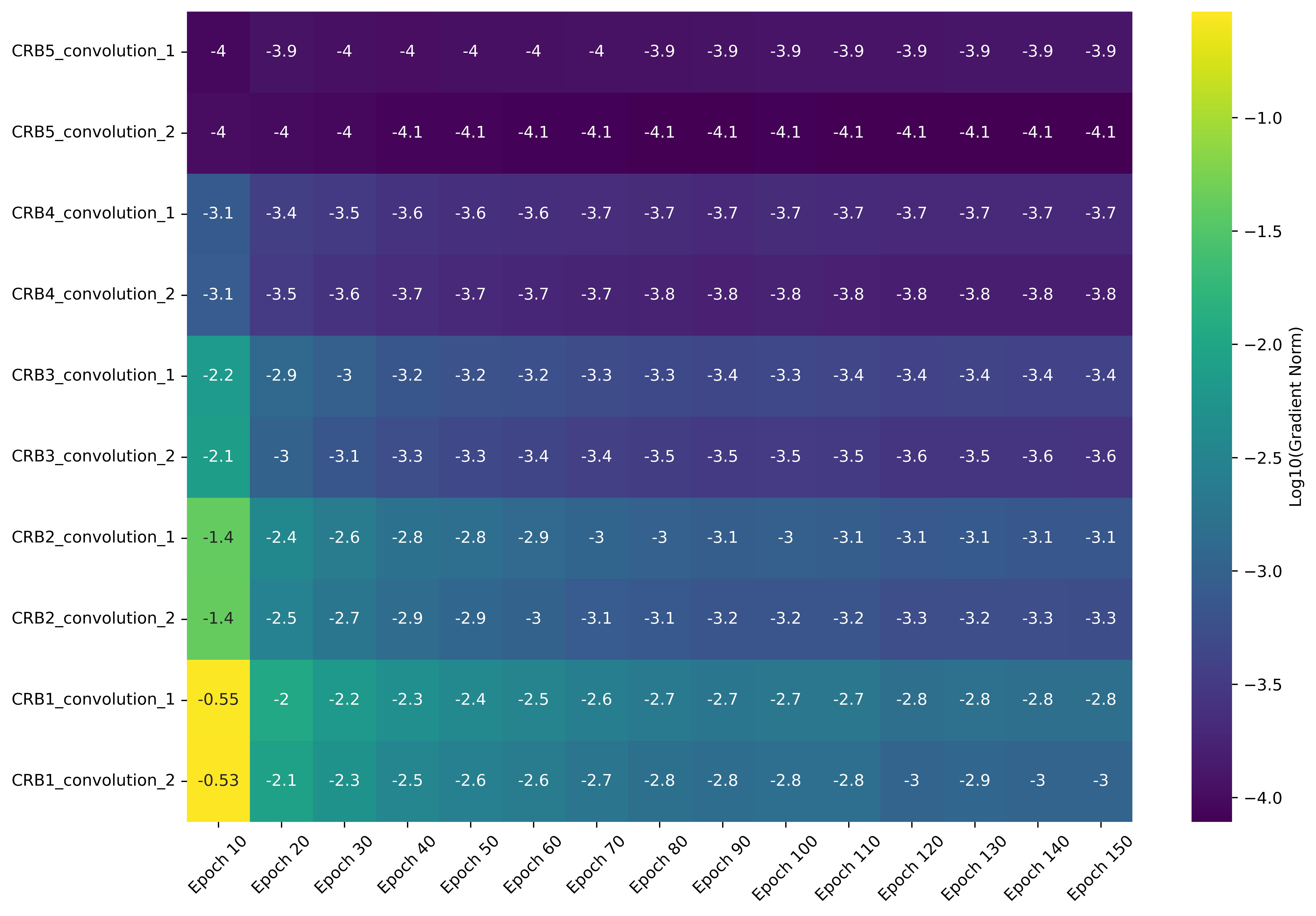}}
	\subfloat[$l=12,\varepsilon=4 (\varepsilon> l/4+\delta_{\text{safe}})$\label{explosion2}]{\includegraphics[width=0.33\textwidth]{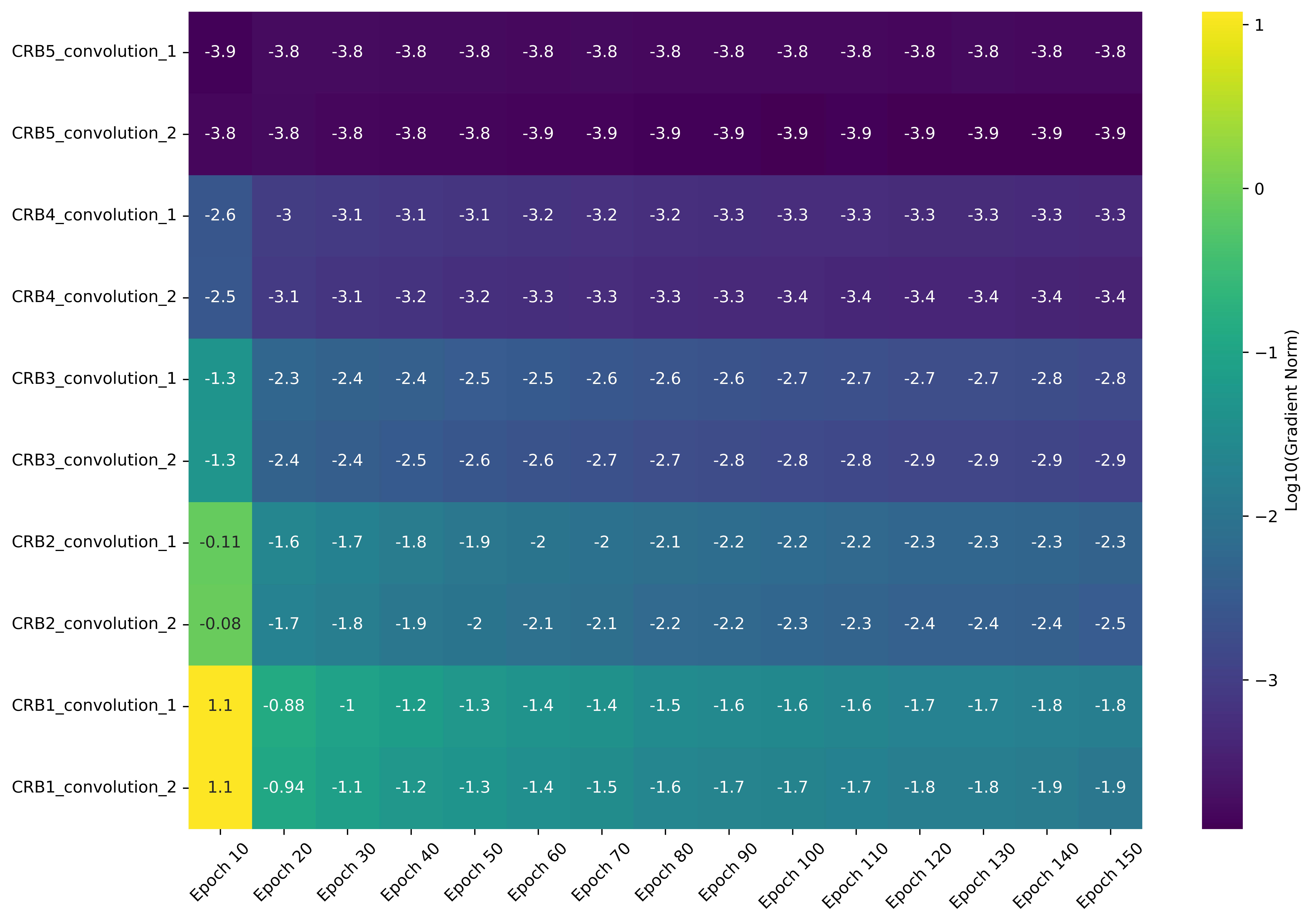}}
	\subfloat[$l=16,\varepsilon=5 (\varepsilon> l/4+\delta_{\text{safe}})$\label{explosion3}]{\includegraphics[width=0.33\textwidth]{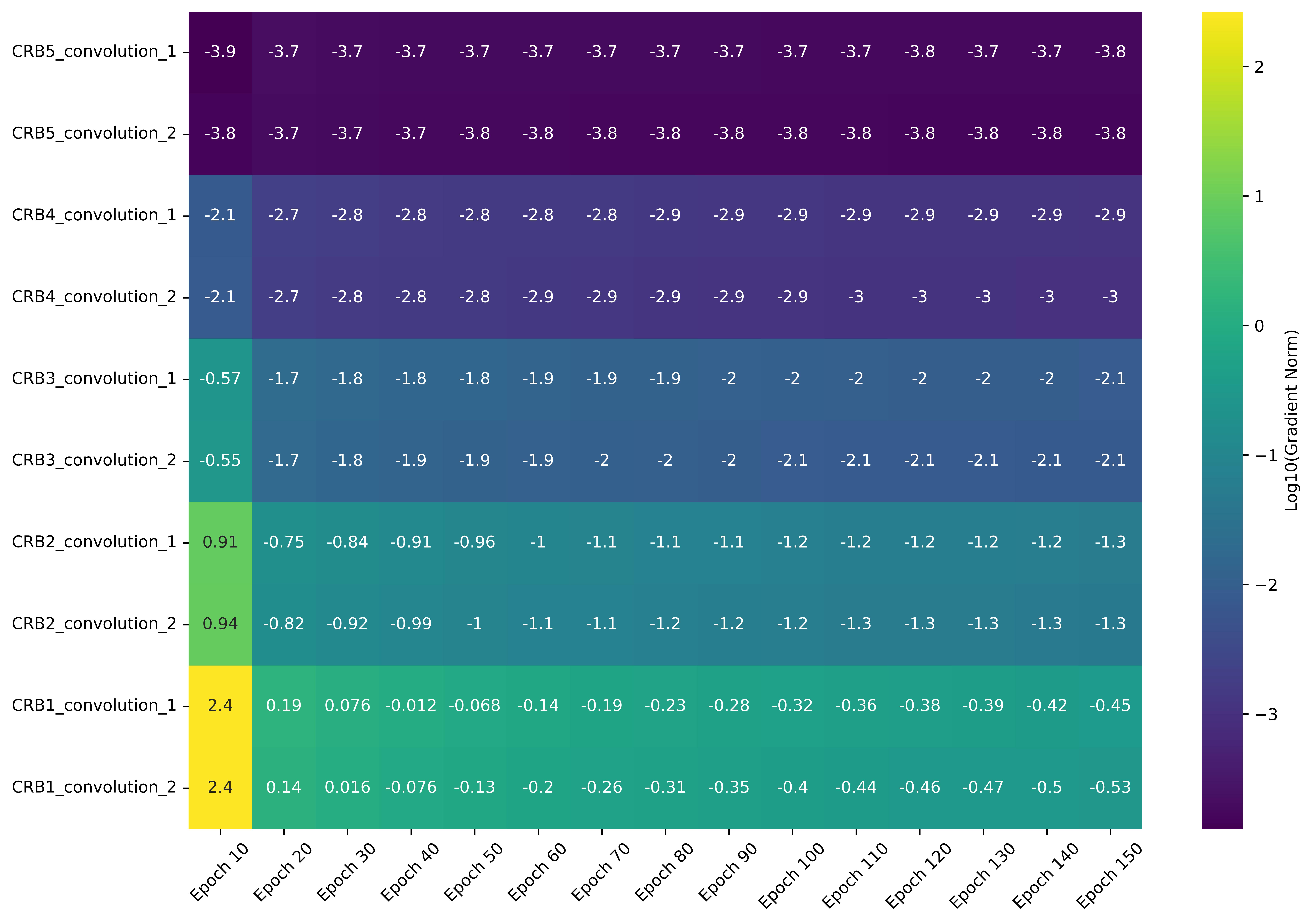}}
	\label{Gradient}
	\caption{Gradient norm heatmaps of CRB variants during the training process. $l$ represents the module depth and $\varepsilon$ the number of residual connections.}
\end{figure*}

Fig.14 reveals a critical phenomenon validating the constraint $l/4+\delta_{\text{safe}}$. For a cascade of $m=5$ CRBs, when $\varepsilon\leqslant l/4+\delta_{\text{safe}}$ is satisfied, the gradient norms $\Vert\nabla_{w}L\Vert$ for the convolution layers within the initial CRB (closest to the input) are smaller than those within the terminal CRB (closest to the output). This is because $J_l^{\text{CRB}_{\varepsilon}}$ for each block satisfies $\Vert J_l^{\text{CRB}_{\varepsilon}}\Vert_2>1$ but still remains bounded, preventing gradient vanishing and explosion. The total gradients at the input of block $k$ can be expressed as:
\begin{equation}
	\Vert\nabla_{x_k}L\Vert_2 \propto \Vert\nabla_{x_m}L\Vert_2\cdot\prod_{j=k}^{m-1}\Vert J_j^{\text{CRB}_{\varepsilon}}\Vert_2
\end{equation}

While the residual structure $\Vert J_j^{\text{CRB}_{\varepsilon}}\Vert_2>1$ mitigates vanishing gradient, the cumulative product over blocks still induces controlled attenuation towards the input, resulting in $\Vert\nabla_{w_1}L\Vert<\Vert\nabla_{w_m}L\Vert$, as shown in Fig.\ref{CRB_8l_2epsilon}, \ref{CRB_12l_3epsilon}, \ref{CRB_16l_4epsilon}. This reflects stable, well-behaved back propagation.

When $\varepsilon>l/4+\delta_{\text{safe}}$, however, the gradient norms $\Vert\nabla_{w}L\Vert$ for convolution layers within the initial CRB becomes larger than those within the terminal CRB. This is because violating $\varepsilon\leqslant l/4+\delta_{\text{safe}}$ triggers $\Vert J_l^{\text{CRB}_{\varepsilon}}\Vert>2$. Therefore, gradient at the input of block $k$ scales exponentially as:
\begin{equation}
	\Vert\nabla_{x_k}L\Vert_2\propto \Vert\nabla_{x_m}L\Vert_2 \cdot \prod_{j=k}^{m-1}\Vert J_j^{\text{CRB}_{\varepsilon}}\Vert_2>\Vert\nabla_{x_m}L\Vert_2 \cdot 2^{(m-k)}
\end{equation}
where the exponential amplification dominates the gradient flow. Given that the input block undergoes amplification over Jacobian multiplications, it is obvious that $\Vert\nabla_{w_1}L\Vert_2>\Vert\nabla_{w_m}L\Vert_2$, manifesting the gradient explosion predicted theoretically. Fig.\ref{explosion1}, \ref{explosion2}, \ref{explosion3} reveal the explosion's origin in the earliest layers in CRB\textsubscript{1}, and this abnormal gradient behaviour directly explains the significant performance degradation observed in Table \ref{ablation epsilon}.

\section{Conclusions}
In this study, we propose the concept of ``Universality'' to quantify the transferability of modules. We introduce the UAE to measure module universality and design two optimized variants, CRB and DCRB, on the basis of RB. Analysis on their back propagation process reveals the underlying mechanisms behind these optimization behaviors. Extensive experiments across multiple SISR scenarios and other low-level tasks demonstrate that our approach outperforms several state‑of‑the‑art models. Moreover, the proposed optimization strategy is applicable to other basic modules, providing a principled framework for future architectural advances.

\section*{Author Contributions}
\textbf{Haotong Cheng:} Conceptualization, formal analysis, methodology, software, writing-original draft. \textbf{Zhiqi Zhang:} Investigation, validation, visualization. \textbf{Hao Li and Xinshang Zhang:} Resources, supervision, writing-review \& editing.

\section*{Conflicts of Interest}
The authors declare no conflicts of interest.

\section*{Data Availability Statement}
This study uses publicly available datasets. The datasets for training are available via the following links: DIV2K(\url{https://data.vision.ee.ethz.ch/cvl/DIV2K/}), BelT(\url{https://github.com/CUMT-AIPR-Lab/CUMT-AIPR-Lab}). Datasets for evaluation are accessible in the following link: Set5, Set14, B100, Urban100(\url{https://gitcode.com/Resource-Bundle-Collection/efef9/}). Other Low-Level datasets could be found directly in their corresponding references. Experimental datas are provided within the article.


\end{document}